\documentclass{article}
\usepackage{ijcai24}
\usepackage{epsfig}
\usepackage{times}
\usepackage{soul}
\usepackage{url}
\usepackage[hidelinks]{hyperref}
\usepackage[utf8]{inputenc}
\usepackage[small]{caption}
\usepackage{graphicx}
\usepackage{amsmath}
\usepackage{amsthm}
\usepackage{booktabs}
\usepackage{algorithm}
\usepackage{algorithmic}
\usepackage[switch]{lineno}
\usepackage{multirow}
\usepackage{bm}
\usepackage[table]{xcolor}
\usepackage{pifont}
\usepackage{makecell}

\hypersetup{
    colorlinks,
    citecolor=cyan,
    linkcolor=red, 
    urlcolor=cyan 
}
\usepackage{amsfonts}

\title{ReplayCAD: Generative Diffusion Replay for Continual Anomaly Detection}

\author{
Lei Hu$^1$
\and
Zhiyong Gan$^{1,2}$\and
Ling Deng$^2$\and
Jinglin Liang$^1$ \and
Lingyu Liang$^1$ \and \\Shuangping Huang$^{1,3,}$\thanks{Corresponding author.}  \and Tianshui Chen$^4$\\
\affiliations
$^1$South China University of Technology\\
$^2$China United Network Communications Corporation Limited Guangdong Branch\\
$^3$Pazhou Laboratory \hspace{2em}
$^4$Guangdong University of Technology\\
\emails
eehulei@mail.scut.edu.cn,
eehsp@scut.edu.cn
}

\begin{document}

\maketitle

\begin{abstract}

Continual Anomaly Detection (CAD) enables anomaly detection models in learning new classes while preserving knowledge of historical classes. CAD faces two key challenges: catastrophic forgetting and segmentation of small anomalous regions. Existing CAD methods store image distributions or patch features to mitigate catastrophic forgetting, but they fail to preserve pixel-level detailed features for accurate segmentation. To overcome this limitation, we propose {\bf ReplayCAD}, a novel diffusion-driven generative replay framework that replay high-quality historical data, thus effectively preserving pixel-level detailed features. Specifically, we compress historical data by searching for a class semantic embedding in the conditional space of the pre-trained diffusion model, which can guide the model to replay data with fine-grained pixel details, thus improving the segmentation performance. However, relying solely on semantic features results in limited spatial diversity. Hence, we further use spatial features to guide data compression, achieving precise control of sample space, thereby generating more diverse data. Our method achieves state-of-the-art performance in both classification and segmentation, with notable improvements in segmentation: 11.5\% on VisA and 8.1\% on MVTec. Our source code is available at \url{https://github.com/HULEI7/ReplayCAD}.

\end{abstract}

\section{Introduction}


\begin{figure}[t]
\centerline{\epsfig{figure=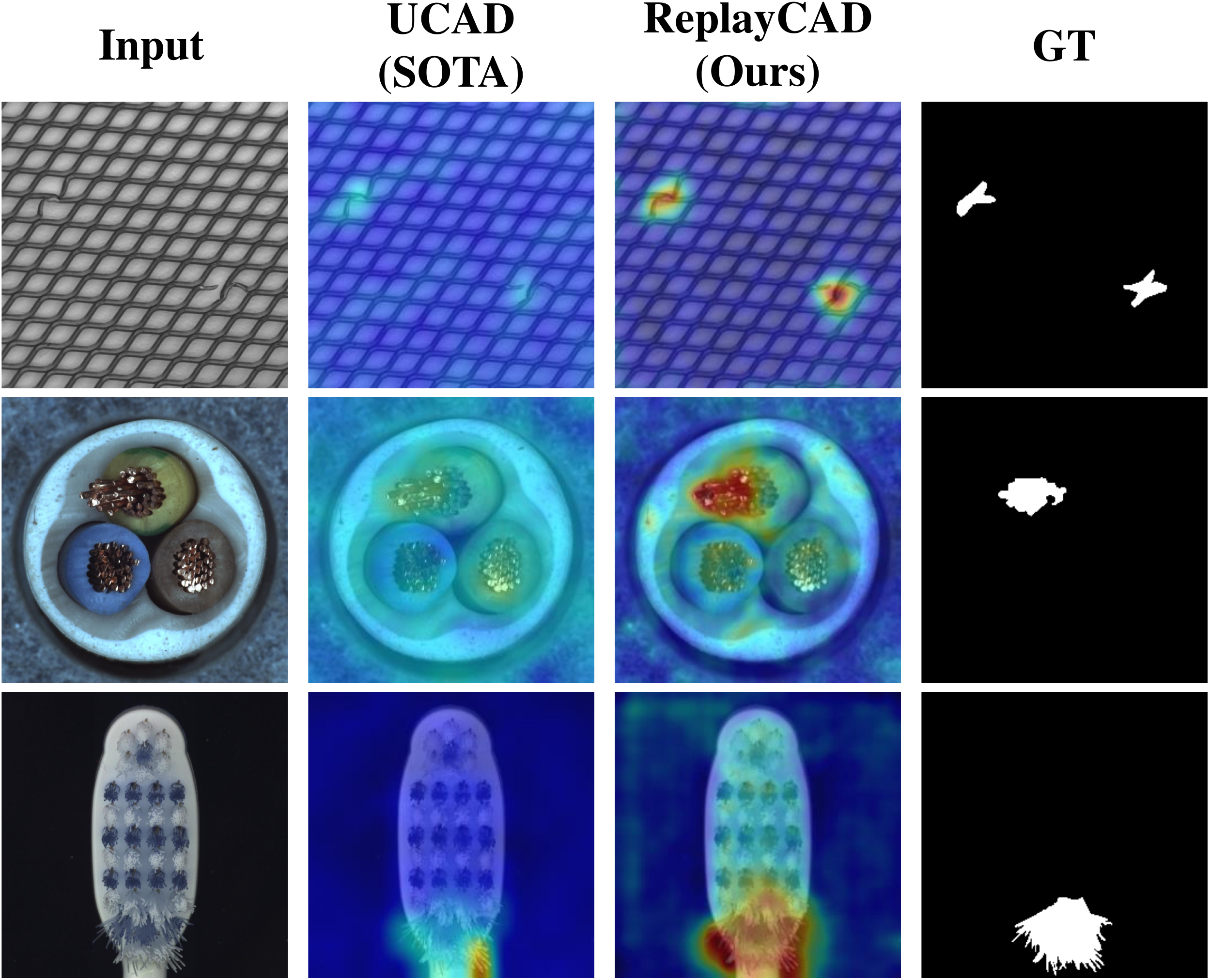,width=8cm}}
\caption{Compared to the SOTA method UCAD, our approach achieves more accurate anomaly segmentation, since we better preserves pixel-level detailed features through high-quality data replay.}
\label{fig:1}
\end{figure}

Nowadays, unsupervised Anomaly Detection (AD) is crucial in modern manufacturing~\cite{industrial_yingyong}. It automatically identifies defects in samples at image and pixel levels, enhancing production efficiency and product quality while reducing reliance on manual inspections. However, in practical industrial production, changes in the generation plan continuously produce new class samples, requiring the model to learn knowledge from new samples while retaining knowledge of previous ones, i.e., Continual Anomaly Detection (CAD).

\begin{figure}[t]
\hspace*{0cm} 
\centerline{\epsfig{figure=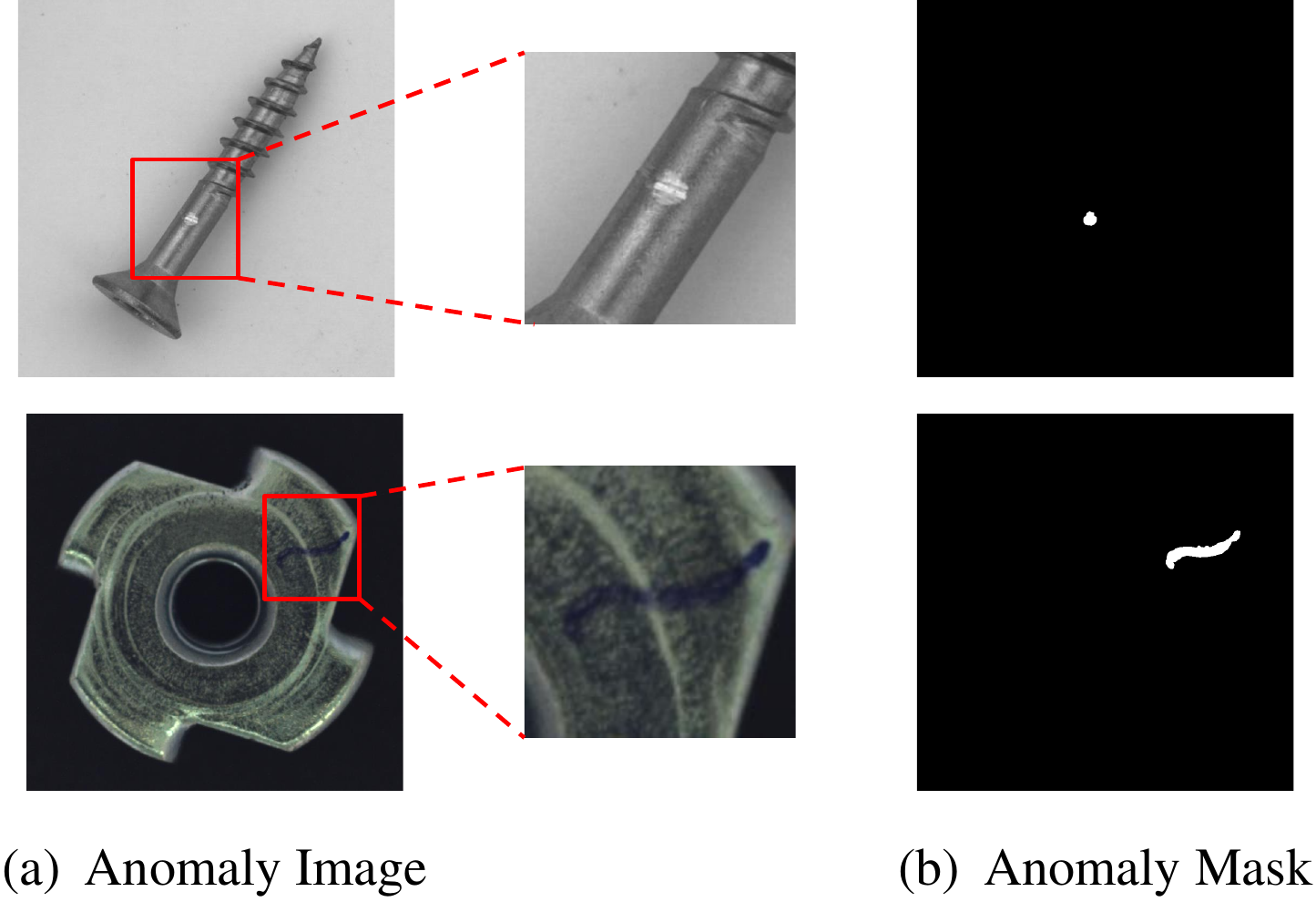,width=7.5cm}}
\caption{Visualization of anomalous regions, illustrating their typically small areas.}
\label{fig:11}
\end{figure}

The key challenge of CAD lies in two aspects: First, catastrophic forgetting, where models tend to catastrophically forget historical class knowledge when learning new class data. Second, CAD requires not only the identification of image-level anomalous samples but also the segmentation of pixel-level anomalous regions. Since anomaly regions are usually small (As illustrated in Figure \ref{fig:11}), this requires the model to remember more detailed pixel-level features from historical samples to ensure accurate anomaly segmentation.

A straightforward approach to addressing these two challenges is to apply existing Continual Learning (CL) methods (e.g. EWC~\cite{Regularization-1}, MAS~\cite{Regularization-3}) to anomaly detection models. However, existing CL methods do not focus on preserving the detailed features of historical samples, making them ineffective for CAD~\cite{IUF}. Advanced CAD methods~\cite{DNE,UCAD} aim to mitigate catastrophic forgetting by compressing and storing features of historical samples. Specifically, DNE~\cite{DNE} stores the mean and variance of image features. However, DNE remembers only the overall features of the distribution while ignoring the detailed features of individual samples. UCAD~\cite{UCAD} divides the image into multiple patches and encodes them into features for storage, resulting in the loss of pixel-level details from historical samples. In summary, these methods fail to preserve the pixel-level detailed features of samples, which negatively impacts segmentation performance.

\begin{figure}[t]
\hspace*{-0.5cm} 
\centerline{\epsfig{figure=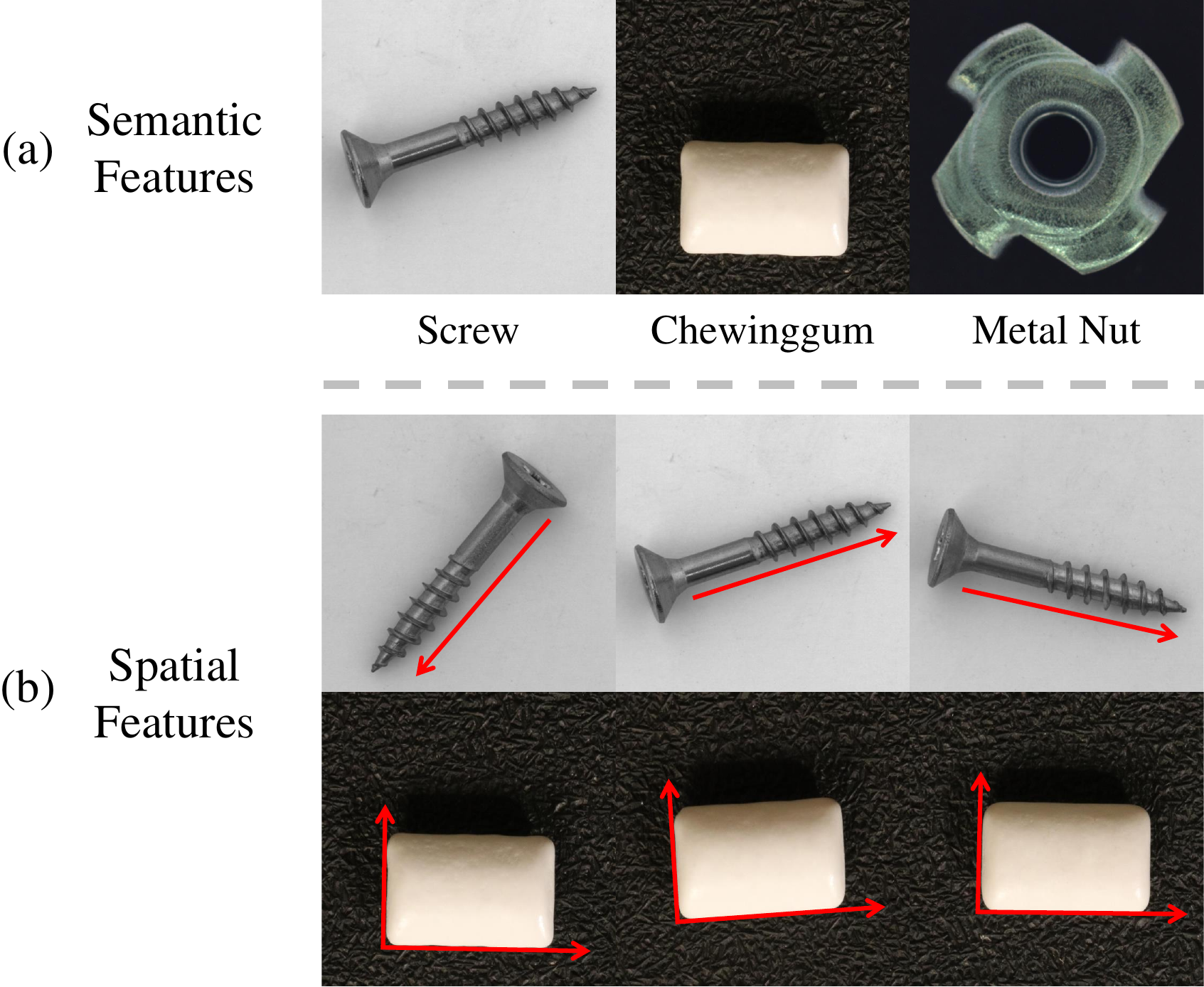,width=7.5cm}}
\caption{Visualization of semantic and spatial features. (a) Semantic features, samples from different classes exhibit distinct texture and shape characteristics. (b) Spatial features, samples within the same class are distributed across different spatial locations.}
\label{fig:2}
\end{figure}

To address this, we propose {\bf ReplayCAD}, a novel CAD framework that can replay real and diverse historical data, thus preserving pixel-level detailed features. ReplayCAD consists of two stages: (1) compressing historical data of each class into conditional features; and (2) utilizing these features to replay historical data, and using the replayed data together with new class data for model training. Our key contribution lies in enhancing first-stage data compression to generate more diverse and authentic historical data. Specifically, in the first stage, we employ reverse engineering to learn a class semantic embedding in the conditional space of a pre-trained diffusion model for each class's historical data. This class semantic embedding guides the diffusion model to generate historical data with pixel-level detailed features, significantly enhancing the segmentation performance of anomalous regions. However, as illustrated in Figure \ref{fig:2}, samples contain not only semantic features but also rich spatial features. Compressing data using semantic features alone leads to replayed data lacking spatial diversity. To address this issue, we utilize the sample's mask as its spatial features to guide the data compression process. This approach enables controlled generation of spatial positions, resulting in the replay of more diverse historical data. Our method offers several advantages: Firstly, compared to current CAD methods based on feature replay~\cite{DNE,UCAD} and regularization~\cite{IUF,CDAD}, the replay-based approach is more effective in preserving the detailed features of samples, which leads to better performance in anomaly segmentation. Secondly, by considering both semantic and spatial features, we generate high-quality and diverse samples, which significantly mitigates catastrophic forgetting. Finally, by guiding the pre-trained diffusion model to perform reverse engineering, we significantly reduce training time and cost compared to existing replay methods~\cite{generate_replay_1,generate_replay_2,generate_replay_3,generate_replay_4} that require training the entire generator.

To sum up, our contributions are as follows:






\begin{itemize}

    \item We propose a novel CAD framework ReplayCAD, which effectively preserves fine-grained pixel details through high-quality data replay, significantly enhancing anomaly segmentation performance. To our best knowledge, this is the first attempt to apply diffusion-driven generative replay approach in CAD.

    \item To generate more diverse historical data, we employ spatial features of samples to guide data compression, thereby achieving spatially controlled data replay.

    \item Extensive experiments demonstrate that our method significantly outperforms SOTA approaches, particularly in segmentation performance, achieving improvements of 11.5\% and 8.1\% on VisA and MVTec, respectively.

\end{itemize}
\section{Related Work}

\subsection{Unsupervised Anomaly Detection}

Currently, unsupervised anomaly detection methods can be classified into the following three types: {\bf Embedding-based methods}~\cite{Embedding_Patchcore,Embedding_RD,Embedding_Simplenet} model the feature distribution of normal samples and detect anomalies by identifying deviations of anomalous samples from this distribution. {\bf Reconstruction-based methods}~\cite{Reconstruction_Uniad,Reconstruction_MambaAD,Reconstruction_Invad,wu2024unsupervised} train a model to reconstruct normal images and detect anomalies by comparing the differences between the input image and the reconstructed image. {\bf Synthesizing-based methods}~\cite{synthesizing_anomalydiffusion,synthesizing_cutpaste,synthesizing_draem,zuo2024clip} train a model by adding artificially synthesized anomalies to normal images, thereby improving anomaly detection capability. However, in real-world industrial scenarios where new classes of data frequently emerge, and existing AD methods lack the capability for continual learning, our goal is to equip existing anomaly detection methods with continual learning capability.


\begin{figure*}[t]
\centerline{\epsfig{figure=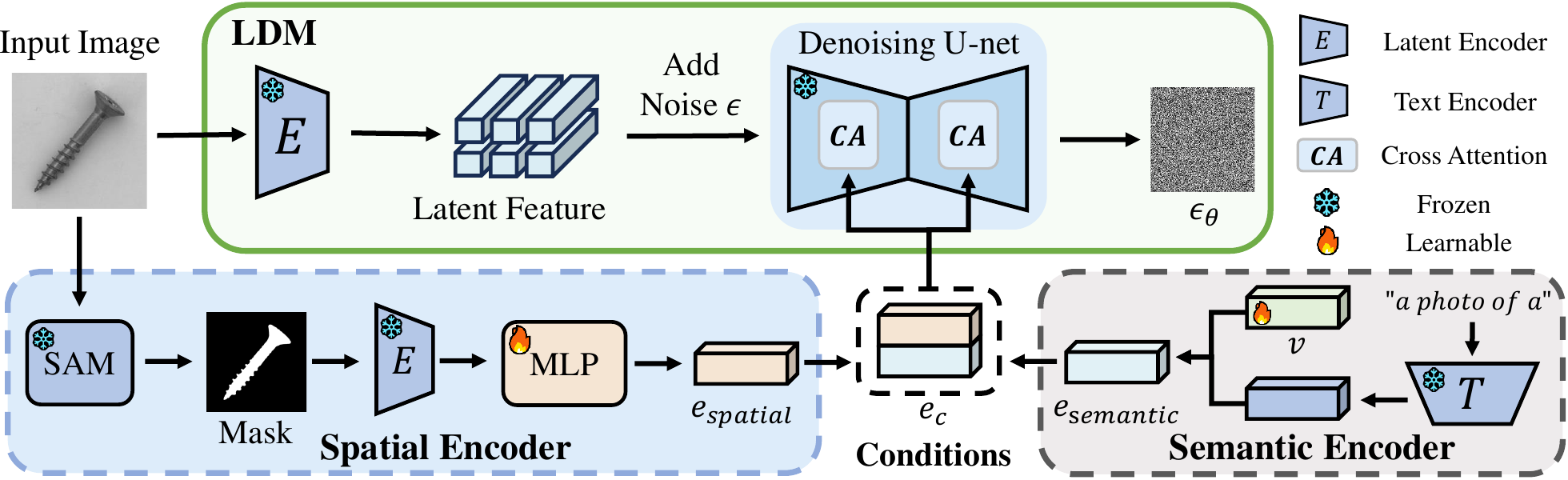,width=16.5cm}}
\caption{Overview of {\bf Data Compression}. We compress the data from both semantic and spatial dimensions: (1) {\bf Semantic}. In the semantic encoder, we get the semantic feature by initializing a class semantic embedding $v$ and combining it with a prompt $p$ encoded by the text encoder. (2) {\bf Spatial}. In the spatial encoder, we first obtain the mask of the sample using SAM and learn the mapping of mask to spatial features by optimizing a MLP layer. Finally, we use spatial and semantic features as conditions to guide the denoising process of LDM. }

\label{fig:3}
\end{figure*}

\subsection{Continual Learning}

The current methods for continual learning can be broadly categorized into the following three types: {\bf Regularization-based methods}~\cite{Regularization-1,Regularization-2,Regularization-3} prevent forgetting of previous tasks by imposing constraints on model parameters. {\bf Replay-based methods}~\cite{Replay-1,Replay-2,Replay-3} consolidate the retention of prior knowledge by replaying data or features from past tasks. {\bf Architecture-based methods}~\cite{Architecture-1,Architecture-2,Architecture-3} assign specific parameters to each task, isolating them to minimize interference. However, existing CL methods are not specifically designed to preserve detailed features of historical samples in CAD, making them unsuitable for direct application to AD models.


\subsection{Continual Anomaly Detection}

Current CAD research can be broadly categorized into two approaches: {\bf Feature replay-based}~\cite{DNE,UCAD} and {\bf Regularization-based}~\cite{IUF,CDAD}. DNE~\cite{DNE} and UCAD~\cite{UCAD} store the feature embeddings of samples from previous classes and locate anomalies by comparing the test sample features with the stored embeddings. However, these methods cannot save the pixel-level detailed features, resulting in poor performance of anomalous region segmentation. IUF~\cite{IUF} and CDAD~\cite{CDAD}prevent knowledge forgetting by restricting parameter updates for previous classes, but its performance deteriorates in long class sequences. To address these issues, we focus on the data itself to remember detailed features, we propose a novel CAD framework that can replay high-quality historical data.

\section{Problem Definition}

In this section, we provide a detailed introduction to the specific setting of the CAD. Unlike traditional AD task where all classes are known during training, CAD requires the model to continuously learn new classes, essentially falling under the category of class-incremental learning. Specifically, for an anomaly detection dataset with $ \mathcal{N} $ classes, its training and testing sets are represented as $\left\{\begin{matrix}{{\mathcal{X}}_{1},{\mathcal{X}}_{2},\cdots ,{\mathcal{X}}_{\mathcal{N}}}\end{matrix}\right\}$ and $\left\{\begin{matrix}{{\mathcal{Y}}_{1},{\mathcal{Y}}_{2},\cdots ,{\mathcal{Y}}_{\mathcal{N}}}\end{matrix}\right\}$, respectively. When learning the $ i$-th class, we can only access the training data $\mathcal{X}_{i}$ and cannot access data from previous classes $\left\{\begin{matrix}{{\mathcal{X}}_{1},{\mathcal{X}}_{2},\cdots ,{\mathcal{X}}_{i-1}}\end{matrix}\right\}$, during testing, we need to perform testing on $\left\{\begin{matrix}{{\mathcal{X}}_{1},{\mathcal{X}}_{2},\cdots ,{\mathcal{X}}_{i}}\end{matrix}\right\}$. Our goal is to sequentially train on the training set of each class and ultimately evaluate on the testing set $\left\{\begin{matrix}{{\mathcal{Y}}_{1},{\mathcal{Y}}_{2},\cdots ,{\mathcal{Y}}_{\mathcal{N}}}\end{matrix}\right\}$.

\section{Methods}

\subsection{Overview}

ReplayCAD consists of two stages: feature-guided data compression stage and replay-enhanced anomaly detection stage. In the first stage, we compress the data of historical classes from both semantic and spatial dimensions, thereby extracting corresponding conditional features, as shown in Figure \ref{fig:3}. Specifically, for semantic features, we search for a class semantic embedding within the conditional space of the pre-trained LDM to guide the model in generating samples for the corresponding class. For spatial features, we learn the mapping relationship between the mask and the spatial features, where the mask is obtained through SAM~\cite{SAM}. In the second stage, we use the conditional features learned in the first stage to replay the historical class data, and train the anomaly detection model together with the new class data. The main contribution of this paper is to enhance the data compression of the first stage from both semantic and spatial dimensions, so as to generate more diversified and authentic historical data. Therefore, in the following, we will focus on how to use semantic and spatial features to guide LDM to compress samples in the first stage.

\subsection{Semantic-Aware Generative Replay}

In this section, we will investigate how to accomplish the two stages of generative replay using only semantic features. 

Latent Diffusion Model (LDM)~\cite{LDM} generates data by denoising in a low-dimensional latent space. Compared to directly working in high-dimensional pixel space, this approach significantly improves computational efficiency while preserving essential semantic information, ensuring high-quality outputs. To further enhance controllability, LDM incorporates a conditional guidance mechanism. This allows conditions such as text descriptions, image masks, or class labels to guide the generation process, ensuring that the results closely align with the given conditions while maintaining diversity and flexibility. Specifically, for the input image $ {x}$ , the LDM first encodes it using a latent encoder $\mathcal{E}$ and then adds random noise $\epsilon$ to generate the noisy image ${{z}_{t}}$ for denoising. Subsequently, the noisy image ${{z}_{t}}$ and conditions $c$ are fed into the denoising U-Net, where the noise ${\epsilon }_{\theta }$ is predicted using a cross-attention mechanism~\cite{SDT,One-dm,peng2024globally,BGTR}. The optimization function of LDM is as follows:

\begin{eqnarray}
{\mathcal{L}}_{LDM} = {\mathbb{E}}_{\mathcal{E}(x),\epsilon \sim \mathcal{N}(0,1),t,c}[\|\epsilon -{\epsilon }_{\theta }(\mathcal{E}(x),t,{\mathcal{E}}_{c}(c))     \|_2^2
]  \text{,}
\end{eqnarray}

\vspace{1mm}

Instead of directly training the LDM, we learn the conditional feature to guide it in generating samples for the corresponding class. Specifically, to represent semantic features, we first randomly initialize a learnable embedding $ {v \in {\mathbb{R}}^{K \times C}}$, which serves as the foundational representation. In addition, we take a text prompt $p$ (e.g., ``\textit{a photo of a}''), encode it using a frozen text encoder $\mathcal{T}$, and then concatenate it with the semantic embedding $ {v}$ to form the final semantic features ${{e}_{semantic}}$, guiding the LDM's denoising process. The optimization objective is as follows:

\vspace{-3mm}

\begin{eqnarray}
        {e}_{semantic} = \{\mathcal{T}(p),{v}\} \text{,}
\end{eqnarray}

\vspace{-3mm}

\begin{eqnarray}
      {v}^{*} = \arg \min_{v}{\mathcal{L}_{LDM}(x,t,{{e}_{semantic}})} \text{,}
\end{eqnarray}

We store the learned semantic embedding $v$ for each class. When new class data appears, we use these semantic embeddings to guide LDM to replay the historical class data. After replaying data from historical classes, we integrate this data with new class data to train the AD model, effectively mitigating the catastrophic forgetting of the historical classes. Our approach imposes no restrictions on anomaly detection models, in this study, we select the multi-class anomaly detection model InvAD~\cite{Invad} as the baseline. The multi-class AD model combines samples from all classes during training, eliminating the need to know sample classes beforehand during inference, thus addressing class uncertainty in CAD. InvAD detects multiple types of anomalies by introducing a feature inversion mechanism and leveraging high-quality feature reconstruction capabilities. Specifically, InvAD consists of an encoder $\textbf{\textit{E}}$ and a generator $\textbf{\textit{G}}$. For a given normal input sample $\textbf{\textit{x}}$, its original feature $\textbf{\textit{F}}$ is first extracted using a frozen encoder, and then its reconstructed feature $\textbf{\textit{F}}^*$ are generated by the generator:

\begin{eqnarray}
      \textbf{\textit{F}} = \textbf{\textit{E}} 
 (\textbf{\textit{x}}) \text{,           }  \textbf{\textit{F}}^* = \textbf{\textit{G}} 
 (\textbf{\textit{F}}) \text{.}
\end{eqnarray}

Then, the abnormal region is obtained by calculating the difference between the original feature $\textbf{\textit{F}}$ and the generated feature $\textbf{\textit{F}}^*$. The loss function is defined as follows:

\begin{eqnarray}
{\mathcal{L}}_{InvAD} = {\mathbb{E}_{ \textbf{\textit{E}},\textbf{\textit{G}},\textbf{\textit{x}}}}[\|\textbf{\textit{F}} -\textbf{\textit{F}}^*    \|_2^2
]  \text{,}
\end{eqnarray} 

\begin{figure}[t]
\centerline{\epsfig{figure=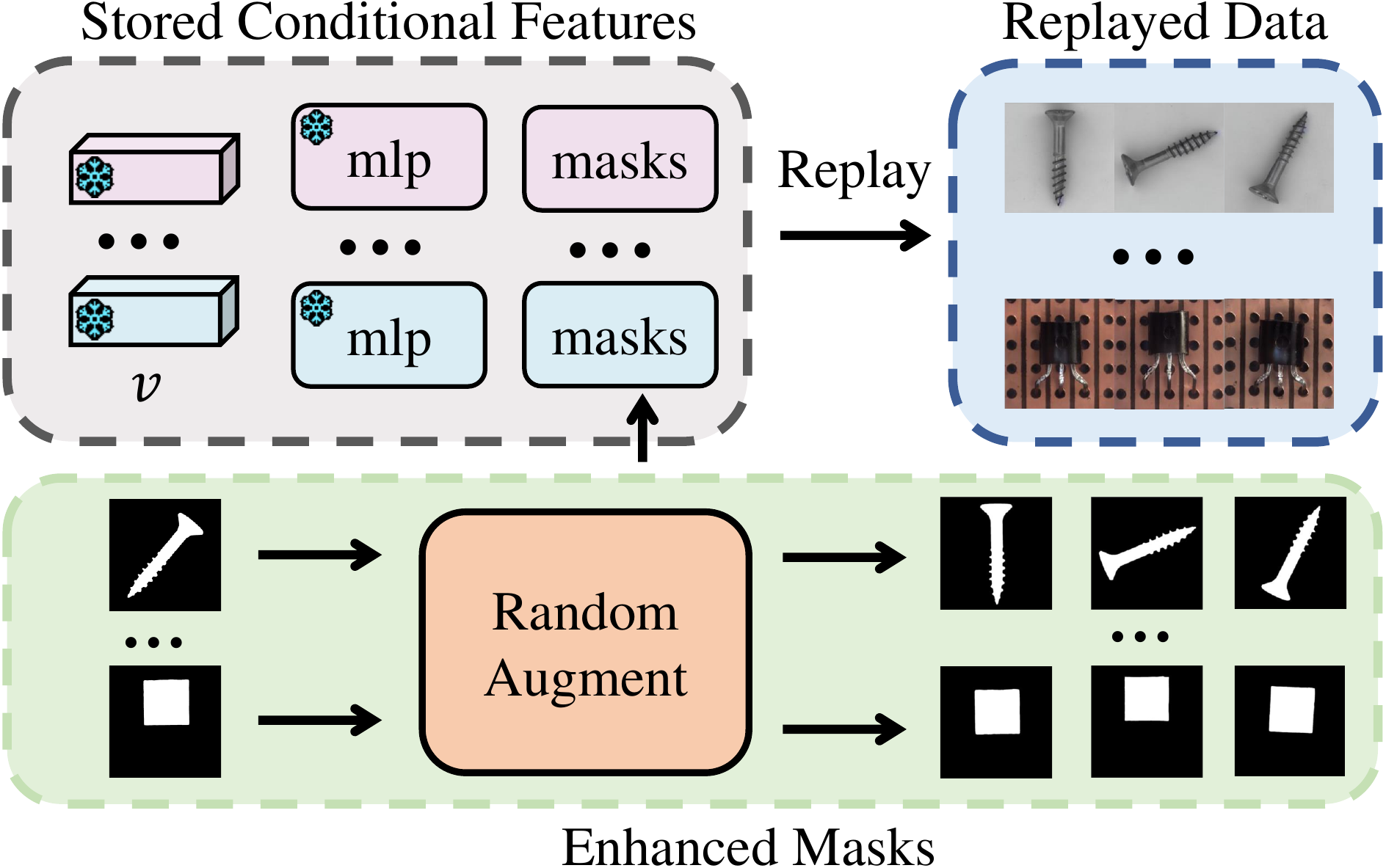,width=8cm}}
\caption{Overview of {\bf Data Replay}. We use stored semantic embeddings $v$, the mlp layer, and mask images to guide the LDM in replaying data from historical classes. To enhance the diversity of replay samples, we apply random augmentations to the masks. }
\label{fig:4}
\end{figure}

\subsection{Spatially-Enhanced Diverse Generation}

However, the samples not only present semantic features, but also various spatial features, as shown in Figure \ref{fig:2}, focusing solely on learning semantic features may cause the generated samples to overfit certain specific examples~\cite{yang2025defect,synthesizing_anomalydiffusion,DDDR}, resulting in a lack of spatial diversity. Therefore, we introduce spatial features to guide the data compression process.

The spatial features of a sample include its coordinates and angle, which are difficult to represent directly. To address this, we use a mask image to denote the sample's spatial features. However, due to the lack of mask labels in the training set, we utilize SAM's ~\cite{SAM} powerful zero-shot segmentation capability to generate the mask for each sample. As the diffusion model cannot directly interpret mask information, we optimize a mlp layer to learn the mapping from the mask to spatial features, thus guiding the diffusion model in denoising. Specifically, for an input image $ {x} \in {\mathbb{R}}^{H \times W \times 3}$, we first generate its corresponding object mask image by SAM. The mask image is then encoded by a latent encoder $\mathcal{E}$ to produce feature representations $ {{x}_{mask}} \in {\mathbb{R}}^{\frac{H}{8} \times \frac{W}{8} \times 4}$, which are then reshaped to match the input size required by the mlp layer, resulting in $ {{x}_{mask}^{'}} \in {\mathbb{R}}^{\frac{HW}{16n} \times n}$. Subsequently, a mlp layer $\mathcal{W}$ with a size of $(n,m)$ is applied to compute the spatial feature $ {{x}_{spatial}} \in {\mathbb{R}}^{\frac{HW}{16n} \times m}$. Finally, $ {{x}_{spatial}}$ is reshaped again to align with the size of the semantic features, yielding the final spatial features $ {{e}_{spatial}} \in {\mathbb{R}}^{M \times C}$.

We concatenate the extracted semantic features ${e}_{semantic}$ and spatial features ${e}_{spatial}$ to obtain the condition ${e}_{c}$, which is used to guide the denoising process of the LDM:

\begin{eqnarray}
  {e}_{c} = \{{e}_{semantic},{e}_{spatial}\}\in {\mathbb{R}}^{(K+M) \times C} \text{,}
\end{eqnarray}

Our goal is to find the best semantic embedding $v$ as well as the mlp layer $\mathcal{W}$ for each class, the optimization objectives are outlined as follows: 

\begin{eqnarray}
      {v}^{*},{\mathcal{W}}^{*} = \arg \min_{v, \mathcal{W} }{\mathcal{L}_{LDM}(x,t,{{e}_{c}})} \text{,}
\end{eqnarray}

\begin{table*}[]
\begin{center}
\tabcolsep=0.2cm
\renewcommand\arraystretch{1.2}
\caption{Comparison with state-of-the-art methods. \textbf{Bold} indicates the best performance, while \underline{underline} indicates the second-best performance. The term " Avg " refers to the mean value obtained by testing across all classes after completing training on the final class.}
\label{tab:1}
\begin{tabular}{cccccccccc}
\specialrule{1.5pt}{0pt}{0pt}
\multirow{3}{*}{\textbf{Method}} & \multirow{3}{*}{\textbf{Venue}} & \multicolumn{4}{c}{\textbf{VisA}} & \multicolumn{4}{c}{\textbf{MVTec}} \\ \cline{3-10} 
 &  & \multicolumn{2}{c}{Image-AUROC} & \multicolumn{2}{c}{Pixel-AP} & \multicolumn{2}{c}{Image-AUROC} & \multicolumn{2}{c}{Pixel-AP} \\ \cline{3-10} 
 &  & Avg (↑) & FM (↓) & Avg (↑) & FM (↓)  & Avg (↑) & FM (↓)  & Avg (↑) & FM (↓)  \\ \hline
 PatchCore & CVPR 22 & 57.6 & 38.0 & 10.3 & 32.4 & 60.7 & 40.7 & 20.9 & 43.1 \\ 
 SimpleNet & CVPR 23 & 58.9 & 33.4 & 7.3 & 30.2 & 62.4 & 36.9 & 17.7 & 32.6 \\
 MambaAD & NeurIPS 24 & 59.5 & 32.4 & 8.8 & 27.6 & 60.6 & 37.6 & 12.7 & 43.8 \\  
 InvAD & arXiv 24 & 67.5 & 28.1 & 9.8 & 33.1 & 66.8 & 31.7 & 21.5 & 38.2 \\ 
 \hline
 SimpleNet+EWC&  CVPR 23  & 59.6 & 35.3 & 9.0 & 28.2 & 69.0 & 29.8 & 21.6 & 28.1 \\
 MambaAD+EWC& NeurIPS 24 & 60.6 & 32.0 & 8.2 & 29.3 & 65.1 & 33.0 & 13.5 & 43.4 \\ 
 SimpleNet+MAS&  CVPR 23   & 60.6 & 34.4 & 9.3 & 28.1 & 70.3 & 28.8 & 25.4 & 24.3 \\ 
 MambaAD+MAS& NeurIPS 24  & 65.2 & 25.9 & 9.5 & 25.6 & 64.5 & 33.9 & 12.2 & 45.2 \\ \hline
 DNE & MM 22 & 60.1 & 17.8 & -- & -- & 87.6 & \underline{4.7} & -- & -- \\ 
 DNE + PANDA & CVPR 21 & 46.0 & 36.8 & -- & -- & 66.3 & 21.8 & -- & -- \\ 
 DNE + CutPaste & CVPR 21 & 61.3 & 18.9 & -- & -- & 83.9 & 9.8 & -- & -- \\ 
  UCAD & AAAI 24& \underline{87.4} & -- & \underline{30.0} & --  & \underline{93.0} & -- & \underline{45.6} & -- \\
 IUF& ECCV 24 & 68.1 & \underline{8.5} & 3.4 & \textbf{0.3} & 76.2 & 6.7 & 17.1 & \underline{5.9} \\ 
CDAD & CVPR 25& 62.8 & 22.2 & 8.3 & 21.7  & 74.3 & 20.1 & 28.0 & 26.1 \\ 
 \hline   \rowcolor{gray!20}
 {\bf ReplayCAD} & \textbf{Ours} & \textbf{90.3} & \textbf{5.5} & \textbf{41.5} & \underline{5.0} & \textbf{94.8} & \textbf{4.5} & \textbf{53.7} & \textbf{5.5} \\ \specialrule{1.5pt}{0pt}{0pt}
\end{tabular}
\end{center}
\end{table*}

After training is completed for each class, we save the semantic embedding $v$ and the mlp layer $\mathcal{W}$ for subsequent data replay. In addition, $M (M \in \left  [1,10\right ])$ randomly selected mask images are stored to control the spatial positioning of the generated images. These stored components constitute the conditional features for each class, requiring minimal storage space while enabling precise control over both semantic and spatial aspects during data replay.

In the second stage, to achieve real and diverse data replay, we use stored conditional features to achieve controllable generation in both semantic and spatial aspects, as illustrated in Figure~\ref{fig:4}. Semantically, we use semantic embedding $v$ to control the class of generated samples. Spatially, we control the spatial position of samples through image masks and the mlp layer $\mathcal{W}$ in the spatial encoder. In addition, we randomly rotate and shift the stored masks to enhance their spatial diversity. By achieving semantic and spatial controllable generation, we can generate real and diverse historical data, thus preserving pixel-level details more efficiently.

\section{Experiments}

\subsection{Experiments Setup}

{\bf Datasets.}  We conducted experiments on two commonly used industrial anomaly detection datasets: VisA~\cite{Visa} and MVTec~\cite{Mvtec}. VisA comprises 12 sample classes, with 8659 training samples (all normal) and 2162 test samples (962 normal, 1200 anomalous).  MVTec includes 15 sample classes, with 3629 training samples (all normal) and 1725 test samples (467 normal, 1258 anomalous).

{\bf Compared Methods.} We compare our approach with three methods: (1) {\bf CAD Methods.} We compare our method with SOTA CAD methods, including CDAD~\cite{CDAD}, IUF~\cite{IUF}, UCAD~\cite{UCAD}, and DNE~\cite{DNE}. Additionally, we migrate the framework of DNE to PANDA~\cite{PANDA} and CutPaste~\cite{synthesizing_cutpaste}. (2) {\bf AD Methods.} We apply current SOTA AD methods directly to CAD, including PatchCore~\cite{Embedding_Patchcore}, SimpleNet~\cite{Embedding_Simplenet}, MambaAD~\cite{Reconstruction_MambaAD} and InvAD~\cite{Invad}, we fine-tune the old model as new class data comes in. (3) {\bf AD+CL Methods.} We adapt CL methods, such as EWC~\cite{Regularization-1} and MAS~\cite{Regularization-3}, to SimpleNet and MambaAD to enhance their adaptability in CAD.

\begin{figure*}[t]
\centerline{\epsfig{figure=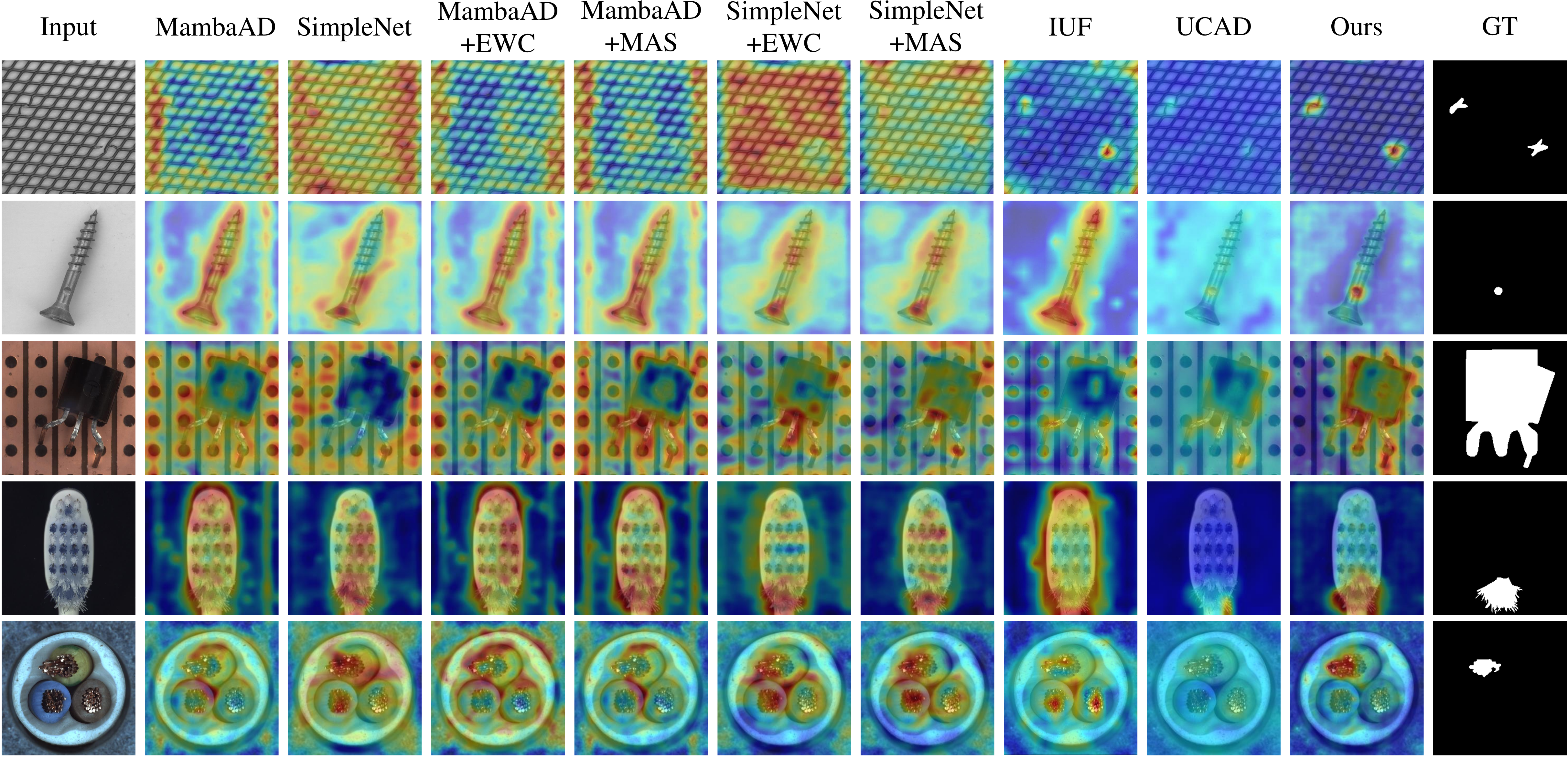,width=17.5cm}}
\caption{Visualization of anomalous region segmentation. From left to right are the test image, the heatmaps predicted by the comparison methods, the heatmap predicted by our method, and the Ground Truth (GT). Areas with high anomaly scores are shown in red, while areas with low anomaly scores are shown in blue.}
\label{fig:5}
\end{figure*}

{\bf Metrics.} Similar to previous methods, we assess the performance of anomaly detection from both classification and segmentation perspectives. For classification performance, we use Area Under the Receiver Operating Characteristics (AUROC)~\cite{AUROC}, while for segmentation performance, we use Area Under Precision-Recall (AP)~\cite{AP}. Moreover, we use Forgetting Measure (FM)~\cite{FM} to evaluate the extent to which the model forgets historical classes:

\begin{eqnarray}
\label{eq:9}
  FM = \frac{1}{\mathcal{N}-1}\displaystyle\sum_{i=1}^{\mathcal{N}-1} \max_{\gamma \in \left\{\begin{matrix}{1,\cdots ,\mathcal{N}-1}\end{matrix}\right\}}{({a}_{\gamma,i}-{a}_{\mathcal{N},i})}  \text{,}
\end{eqnarray}
Here, $\mathcal{N}$ represents the total number of classes, $i$ denotes the class number, and $a$ indicates the classification or segmentation score for that class.

{\bf Implementation Details.} We set the number of tokens \( K \) for the semantic embedding $v$ to 20. For ViSA, we use pre-trained Stable Diffusion 1.5 weights with a resolution of (512, 512), where the dimension \( C \) of the semantic and spatial features is 768, and the MLP layer size is (128, 196). For MVTec, we use pre-trained LDM weights with a resolution of (256, 256), where the dimension \( C \) of the semantic and spatial features is 1280, and the MLP layer size is (128, 200). We generate 800 samples for each class for training the anomaly detection model. We use Invad~\cite{Invad} as the baseline model for anomaly detection, keeping all settings consistent with the original paper.

\begin{table}[]
\begin{center}
\tabcolsep=0.3cm
\renewcommand\arraystretch{1.1}
\caption{Comparison of storage space with the SOTA method.}
\label{tab:2}
\begin{tabular}{ccc}
\specialrule{1.pt}{0pt}{0pt}
Method & VisA & MVTec \\ \hline 
UCAD & 18.6 MB & 23.3 MB \\  
CDAD & 2.7 GB & 3.5 GB\\  \rowcolor{gray!20}
ReplayCAD (Ours) & {\bf1.9 MB} & {\bf3.0 MB} \\ \specialrule{1.pt}{0pt}{0pt}
\end{tabular}
\end{center}
\end{table}

\begin{table}[]
\begin{center}
\tabcolsep=0.2cm
\renewcommand\arraystretch{1.1}
\caption{Effect of semantic and spatial features on MVTec. }
\label{tab:3}
\begin{tabular}{c|cc|cc}
\specialrule{1.pt}{0pt}{0pt}
 Model & Semantic & Spatial & I-AUROC & P-AP \\ \hline
 T1 & \ding{55} & \ding{55} & 66.8 & 21.5  \\
 T2 & \ding{51} & \ding{55} & 92.4 & 49.5 \\
 T3 & \ding{55} & \ding{51} & 86.9 & 42.5 \\ 
 T4 & \ding{51} & \ding{51} & {\bf 94.8} & {\bf 53.7} \\ \specialrule{1.pt}{0pt}{0pt}
\end{tabular}
\end{center}
\end{table}

\begin{table}[]
\begin{center}
\tabcolsep=0.3cm
\renewcommand\arraystretch{1.1}
\caption{Transferring to different AD models on MVTec.}
\label{tab:4}
\begin{tabular}{cc|cc}
\specialrule{1.pt}{0pt}{0pt}
 Method & Venue & I-AUROC & P-AP \\ \hline
 PatchCore & CVPR 22 & 93.7 & {\bf 54.7} \\
 RLR & ECCV 24 & 94.0 & 50.6 \\
 MambaAD & NeurIPS 24 & {\bf 95.3} & 53.1 \\
 InvAD & arXiv 24 & 94.8 & 53.7 \\ \specialrule{1.pt}{0pt}{0pt}
\end{tabular}
\end{center}
\end{table}

\subsection{Experiment Performance}

{\bf Quantitative Evaluation.} As shown in Table~\ref{tab:1}, our method achieves state-of-the-art performance on both image-level and pixel-level metrics across two datasets. Specifically, we improved pixel-level metrics by 11.5\% on VisA and 8.1\% on MVTec compared to the previous state-of-the-art method. Existing anomaly detection methods often suffer from catastrophic forgetting in continual learning scenarios. Although some continual learning methods can mitigate this issue, there remains a significant gap compared to our approach. We observe that although IUF performs poorly in terms of pixel-level and image-level metrics, it demonstrates relatively low FM scores. This is because IUF experiences severe feature conflicts between new and historical classes when learning data from new classes. This conflict causes the historical best performance ${a}_{\gamma,i}$ (Eq.~\ref{eq:9}) of new classes to fall below their actual potential, thus resulting in lower FM scores. Additionally, although UCAD reported the FM metric in the original paper, its historical best score ${a}_{\gamma,i}$ was achieved by storing a larger number of patch features, which fails to reflect the true FM metric. Therefore, we marked it as “-”.

\begin{figure*}[t]
\centerline{\epsfig{figure=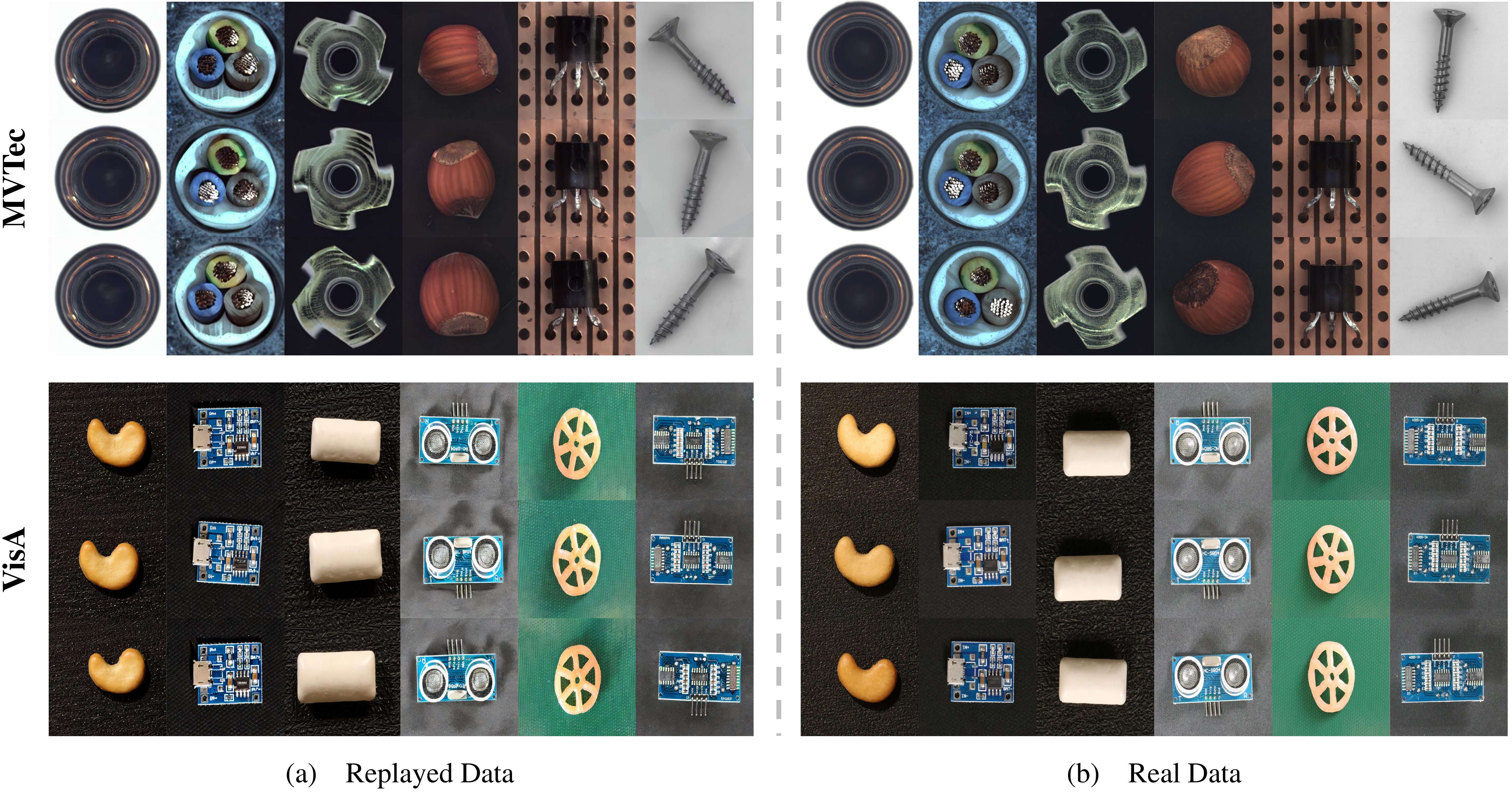,width=17.3cm}}
\caption{The visual comparison of replayed data and real data on the MVTec and VisA datasets.}
\label{fig:6}
\end{figure*}

Additionally, we compare our method's storage space with the SOTA methods. As shown in Table~\ref{tab:2}, different from UCAD storage patch-level features and CDAD storage gradient projection, our approach significantly reduces storage space by storing semantic and spatial features of samples, while also achieving superior experimental performance.

{\bf Qualitative Evaluation.} Figure~\ref{fig:5} shows the heatmaps of anomaly regions predicted by our method and other SOTA methods. We found that the AD methods, AD + CL methods, and IUF often result in a large area of error prediction. This occurs because, when learning new classes, these methods conflict with the features of historical classes, leading to catastrophic forgetting of previously acquired knowledge. Although UCAD shows significant improvement compared to previous methods, it still suffers from missed detections in anomalous regions. Visualization results demonstrate that our method achieves higher accuracy in anomaly segmentation.

{\bf Visualization of Replayed Data.} As shown in Figure~\ref{fig:6}, we present a comparison between the replayed data and the real data. Since AD is conducted in an unsupervised setting, the replayed data should closely resemble the real normal data. The results indicate that the replayed data exhibit a high degree of similarity to the real data, validating the effectiveness of our framework based on generative replay.

\subsection{Analysis}


In this section, we analyze the impact of the two proposed features and different AD models on MVTec.

{\bf Effect of semantic and spatial features.} As shown in Table~\ref{tab:3}, we analyze the effectiveness of the proposed semantic features ${{e}_{semantic}}$ and spatial features ${{e}_{spatial}}$. {\bf T1} indicates that we do not retain any features of historical samples but instead directly train the AD model sequentially on samples from each class. In this case, the model suffers from catastrophic forgetting, resulting in a sharp decline in performance. {\bf T2} and {\bf T3} represent the processes of learning and storing spatial features and semantic features of the samples, respectively, followed by replaying historical data using the learned features. The results indicate that both types of features can significantly alleviate catastrophic forgetting. {\bf T4} simultaneously learns both spatial and semantic features. Compared to learning them individually, this approach significantly improves classification and segmentation performance, further demonstrating the critical role of these two types of features in replaying historical data.

{\bf Transferring to different AD models.} To validate the generalization capability of the replayed data, we transfer it to various anomaly detection models for training, including Patchcore~\cite{Embedding_Patchcore}, RLR~\cite{RLR}, MambaAD~\cite{Reconstruction_MambaAD}, and InvAD~\cite{Invad}, as shown in Table~\ref{tab:4}. Our replayed data demonstrated strong performance across four methods, confirming the superior generalization ability of the proposed method. Moreover, this indicates that our method can be readily adapted to SOTA AD methods at any time. Compared to other CAD methods that can only adapt to a single AD method, our approach demonstrates greater adaptability and sustainability. 

\section{Conclusion}

In this paper, we introduce ReplayCAD, marking the first use of generative replay based on a pre-trained diffusion model in CAD. Compared to current state-of-the-art methods, our approach more effectively preserves the detailed features of historical samples, leading to significant improvements in segmenting anomalous regions. Additionally, by incorporating semantic and spatial feature guidance into data compression, we achieve high-quality data generation. Experimental results demonstrate that our method achieves SOTA performance on both image-level and pixel-level metrics.


\section*{Acknowledgments}
The research is partially supported by National Natural Science Foundation of China (No.62176093, 61673182), National Key Research and Development Program of China (No.2023YFC3502900), Key Realm Research and Development Program of Guangzhou (No.202206030001), Guangdong-Hong Kong-Macao Joint Innovation Project (No.2023A0505030016), Guangdong Emergency Management Science and Technology Program (No.2025YJKY001).

\bibliographystyle{named}
\bibliography{main}

\begin{thebibliography}{}

\bibitem[\protect\citeauthoryear{Aljundi \bgroup \em et al.\egroup }{2018}]{Regularization-3}
Rahaf Aljundi, Francesca Babiloni, Mohamed Elhoseiny, Marcus Rohrbach, and Tinne Tuytelaars.
\newblock Memory aware synapses: Learning what (not) to forget.
\newblock In {\em ECCV}, pages 139--154, 2018.

\bibitem[\protect\citeauthoryear{Antonelli \bgroup \em et al.\egroup }{2022}]{MSD}
Michela Antonelli, Annika Reinke, Spyridon Bakas, Keyvan Farahani, Annette Kopp-Schneider, Bennett~A Landman, Geert Litjens, Bjoern Menze, Olaf Ronneberger, Ronald~M Summers, et~al.
\newblock The medical segmentation decathlon.
\newblock {\em Nature communications}, page 4128, 2022.

\bibitem[\protect\citeauthoryear{Bergmann \bgroup \em et al.\egroup }{2019}]{Mvtec}
Paul Bergmann, Michael Fauser, David Sattlegger, and Carsten Steger.
\newblock Mvtec ad--a comprehensive real-world dataset for unsupervised anomaly detection.
\newblock In {\em CVPR}, pages 9592--9600, 2019.

\bibitem[\protect\citeauthoryear{Chaudhry \bgroup \em et al.\egroup }{2018}]{FM}
Arslan Chaudhry, Puneet~K Dokania, Thalaiyasingam Ajanthan, and Philip~HS Torr.
\newblock Riemannian walk for incremental learning: Understanding forgetting and intransigence.
\newblock In {\em ECCV}, pages 532--547, 2018.

\bibitem[\protect\citeauthoryear{Dai \bgroup \em et al.\egroup }{2023}]{SDT}
Gang Dai, Yifan Zhang, Qingfeng Wang, Qing Du, Zhuliang Yu, Zhuoman Liu, and Shuangping Huang.
\newblock Disentangling writer and character styles for handwriting generation.
\newblock In {\em CVPR}, pages 5977--5986, 2023.

\bibitem[\protect\citeauthoryear{Dai \bgroup \em et al.\egroup }{2024}]{One-dm}
Gang Dai, Yifan Zhang, Quhui Ke, Qiangya Guo, and Shuangping Huang.
\newblock One-dm: One-shot diffusion mimicker for handwritten text generation.
\newblock In {\em ECCV}, pages 410--427. Springer, 2024.

\bibitem[\protect\citeauthoryear{Deng and Li}{2022}]{Embedding_RD}
Hanqiu Deng and Xingyu Li.
\newblock Anomaly detection via reverse distillation from one-class embedding.
\newblock In {\em CVPR}, pages 9737--9746, 2022.

\bibitem[\protect\citeauthoryear{Gao and Liu}{2023}]{generate_replay_3}
Rui Gao and Weiwei Liu.
\newblock Ddgr: Continual learning with deep diffusion-based generative replay.
\newblock In {\em ICML}, pages 10744--10763. PMLR, 2023.

\bibitem[\protect\citeauthoryear{He \bgroup \em et al.\egroup }{2024}]{Reconstruction_MambaAD}
Haoyang He, Yuhu Bai, Jiangning Zhang, Qingdong He, Hongxu Chen, Zhenye Gan, Chengjie Wang, Xiangtai Li, Guanzhong Tian, and Lei Xie.
\newblock Mambaad: Exploring state space models for multi-class unsupervised anomaly detection.
\newblock {\em arXiv preprint arXiv:2404.06564}, 2024.

\bibitem[\protect\citeauthoryear{He \bgroup \em et al.\egroup }{2025}]{RLR}
Liren He, Zhengkai Jiang, Jinlong Peng, Wenbing Zhu, Liang Liu, Qiangang Du, Xiaobin Hu, Mingmin Chi, Yabiao Wang, and Chengjie Wang.
\newblock Learning unified reference representation for unsupervised multi-class anomaly detection.
\newblock In {\em ECCV}, pages 216--232. Springer, 2025.

\bibitem[\protect\citeauthoryear{Heller and others}{2023}]{kits}
Nicholas Heller et~al.
\newblock The kits21 challenge: Automatic segmentation of kidneys, renal tumors, and renal cysts in corticomedullary-phase ct, 2023.

\bibitem[\protect\citeauthoryear{Hu and Huang}{2025}]{BGTR}
Lei Hu and Shuangping Huang.
\newblock Enhancing table structure recognition via bounding box guidance.
\newblock In {\em ICPR}, pages 209--225. Springer, 2025.

\bibitem[\protect\citeauthoryear{Hu \bgroup \em et al.\egroup }{2024}]{synthesizing_anomalydiffusion}
Teng Hu, Jiangning Zhang, Ran Yi, Yuzhen Du, Xu~Chen, Liang Liu, Yabiao Wang, and Chengjie Wang.
\newblock Anomalydiffusion: Few-shot anomaly image generation with diffusion model.
\newblock In {\em AAAI}, volume~38, pages 8526--8534, 2024.

\bibitem[\protect\citeauthoryear{Kirillov \bgroup \em et al.\egroup }{2023}]{SAM}
Alexander Kirillov, Eric Mintun, Nikhila Ravi, Hanzi Mao, Chloe Rolland, Laura Gustafson, Tete Xiao, Spencer Whitehead, Alexander~C Berg, Wan-Yen Lo, et~al.
\newblock Segment anything.
\newblock In {\em ICCV}, pages 4015--4026, 2023.

\bibitem[\protect\citeauthoryear{Kirkpatrick \bgroup \em et al.\egroup }{2017}]{Regularization-1}
James Kirkpatrick, Razvan Pascanu, Neil Rabinowitz, Joel Veness, Guillaume Desjardins, Andrei~A Rusu, Kieran Milan, John Quan, Tiago Ramalho, Agnieszka Grabska-Barwinska, et~al.
\newblock Overcoming catastrophic forgetting in neural networks.
\newblock {\em Proceedings of the national academy of sciences}, 114(13):3521--3526, 2017.

\bibitem[\protect\citeauthoryear{Li \bgroup \em et al.\egroup }{2021}]{synthesizing_cutpaste}
Chun-Liang Li, Kihyuk Sohn, Jinsung Yoon, and Tomas Pfister.
\newblock Cutpaste: Self-supervised learning for anomaly detection and localization.
\newblock In {\em CVPR}, pages 9664--9674, 2021.

\bibitem[\protect\citeauthoryear{Li \bgroup \em et al.\egroup }{2022}]{DNE}
Wujin Li, Jiawei Zhan, Jinbao Wang, Bizhong Xia, Bin-Bin Gao, Jun Liu, Chengjie Wang, and Feng Zheng.
\newblock Towards continual adaptation in industrial anomaly detection.
\newblock In {\em ACMMM}, pages 2871--2880, 2022.

\bibitem[\protect\citeauthoryear{Li \bgroup \em et al.\egroup }{2024}]{generate_replay_4}
Wei Li, Jingyang Zhang, Pheng-Ann Heng, and Lixu Gu.
\newblock Comprehensive generative replay for task-incremental segmentation with concurrent appearance and semantic forgetting.
\newblock In {\em MICCAI}, pages 80--90. Springer, 2024.

\bibitem[\protect\citeauthoryear{Li \bgroup \em et al.\egroup }{2025}]{CDAD}
Xiaofan Li, Xin Tan, Zhuo Chen, Zhizhong Zhang, Ruixin Zhang, Rizen Guo, Guanna Jiang, Yulong Chen, Yanyun Qu, Lizhuang Ma, et~al.
\newblock One-for-more: Continual diffusion model for anomaly detection.
\newblock {\em arXiv preprint arXiv:2502.19848}, 2025.

\bibitem[\protect\citeauthoryear{Liang \bgroup \em et al.\egroup }{2024}]{DDDR}
Jinglin Liang, Jin Zhong, Hanlin Gu, Zhongqi Lu, Xingxing Tang, Gang Dai, Shuangping Huang, Lixin Fan, and Qiang Yang.
\newblock Diffusion-driven data replay: A novel approach to combat forgetting in federated class continual learning.
\newblock In {\em ECCV}, pages 303--319. Springer, 2024.

\bibitem[\protect\citeauthoryear{Liu \bgroup \em et al.\egroup }{2020}]{Replay-2}
Xialei Liu, Chenshen Wu, Mikel Menta, Luis Herranz, Bogdan Raducanu, Andrew~D Bagdanov, Shangling Jui, and Joost~van de~Weijer.
\newblock Generative feature replay for class-incremental learning.
\newblock In {\em CVPRW}, pages 226--227, 2020.

\bibitem[\protect\citeauthoryear{Liu \bgroup \em et al.\egroup }{2023}]{Embedding_Simplenet}
Zhikang Liu, Yiming Zhou, Yuansheng Xu, and Zilei Wang.
\newblock Simplenet: A simple network for image anomaly detection and localization.
\newblock In {\em CVPR}, pages 20402--20411, 2023.

\bibitem[\protect\citeauthoryear{Liu \bgroup \em et al.\egroup }{2024a}]{UCAD}
Jiaqi Liu, Kai Wu, Qiang Nie, Ying Chen, Bin-Bin Gao, Yong Liu, Jinbao Wang, Chengjie Wang, and Feng Zheng.
\newblock Unsupervised continual anomaly detection with contrastively-learned prompt.
\newblock In {\em AAAI}, volume~38, pages 3639--3647, 2024.

\bibitem[\protect\citeauthoryear{Liu \bgroup \em et al.\egroup }{2024b}]{industrial_yingyong}
Jiaqi Liu, Guoyang Xie, Jinbao Wang, Shangnian Li, Chengjie Wang, Feng Zheng, and Yaochu Jin.
\newblock Deep industrial image anomaly detection: A survey.
\newblock {\em Machine Intelligence Research}, 21(1):104--135, 2024.

\bibitem[\protect\citeauthoryear{Mallya and Lazebnik}{2018}]{Architecture-1}
Arun Mallya and Svetlana Lazebnik.
\newblock Packnet: Adding multiple tasks to a single network by iterative pruning.
\newblock In {\em CVPR}, pages 7765--7773, 2018.

\bibitem[\protect\citeauthoryear{Mallya \bgroup \em et al.\egroup }{2018}]{Architecture-2}
Arun Mallya, Dillon Davis, and Svetlana Lazebnik.
\newblock Piggyback: Adapting a single network to multiple tasks by learning to mask weights.
\newblock In {\em ECCV}, pages 67--82, 2018.

\bibitem[\protect\citeauthoryear{Pan \bgroup \em et al.\egroup }{2020}]{Regularization-2}
Pingbo Pan, Siddharth Swaroop, Alexander Immer, Runa Eschenhagen, Richard Turner, and Mohammad Emtiyaz~E Khan.
\newblock Continual deep learning by functional regularisation of memorable past.
\newblock {\em NIPS}, 33:4453--4464, 2020.

\bibitem[\protect\citeauthoryear{Peng \bgroup \em et al.\egroup }{2024}]{peng2024globally}
Wenjie Peng, Hongxiang Huang, Tianshui Chen, Quhui Ke, Gang Dai, and Shuangping Huang.
\newblock Globally correlation-aware hard negative generation.
\newblock {\em IJCV}, pages 1--22, 2024.

\bibitem[\protect\citeauthoryear{Rebuffi \bgroup \em et al.\egroup }{2017}]{Replay-1}
Sylvestre-Alvise Rebuffi, Alexander Kolesnikov, Georg Sperl, and Christoph~H Lampert.
\newblock icarl: Incremental classifier and representation learning.
\newblock In {\em CVPR}, pages 2001--2010, 2017.

\bibitem[\protect\citeauthoryear{Reiss \bgroup \em et al.\egroup }{2021}]{PANDA}
Tal Reiss, Niv Cohen, Liron Bergman, and Yedid Hoshen.
\newblock Panda: Adapting pretrained features for anomaly detection and segmentation.
\newblock In {\em CVPR}, pages 2806--2814, 2021.

\bibitem[\protect\citeauthoryear{Riemer \bgroup \em et al.\egroup }{2019}]{Replay-3}
Matthew Riemer, Tim Klinger, Djallel Bouneffouf, and Michele Franceschini.
\newblock Scalable recollections for continual lifelong learning.
\newblock In {\em AAAI}, volume~33, pages 1352--1359, 2019.

\bibitem[\protect\citeauthoryear{Rombach \bgroup \em et al.\egroup }{2022}]{LDM}
Robin Rombach, Andreas Blattmann, Dominik Lorenz, Patrick Esser, and Bj{\"o}rn Ommer.
\newblock High-resolution image synthesis with latent diffusion models.
\newblock In {\em CVPR}, pages 10684--10695, 2022.

\bibitem[\protect\citeauthoryear{Roth \bgroup \em et al.\egroup }{2022}]{Embedding_Patchcore}
Karsten Roth, Latha Pemula, Joaquin Zepeda, Bernhard Sch{\"o}lkopf, Thomas Brox, and Peter Gehler.
\newblock Towards total recall in industrial anomaly detection.
\newblock In {\em CVPR}, pages 14318--14328, 2022.

\bibitem[\protect\citeauthoryear{Shin \bgroup \em et al.\egroup }{2017}]{generate_replay_1}
Hanul Shin, Jung~Kwon Lee, Jaehong Kim, and Jiwon Kim.
\newblock Continual learning with deep generative replay.
\newblock {\em NIPS}, 30, 2017.

\bibitem[\protect\citeauthoryear{Tang \bgroup \em et al.\egroup }{2025}]{IUF}
Jiaqi Tang, Hao Lu, Xiaogang Xu, Ruizheng Wu, Sixing Hu, Tong Zhang, Tsz~Wa Cheng, Ming Ge, Ying-Cong Chen, and Fugee Tsung.
\newblock An incremental unified framework for small defect inspection.
\newblock In {\em ECCV}, pages 307--324. Springer, 2025.

\bibitem[\protect\citeauthoryear{Wang \bgroup \em et al.\egroup }{2021}]{generate_replay_2}
Liyuan Wang, Kuo Yang, Chongxuan Li, Lanqing Hong, Zhenguo Li, and Jun Zhu.
\newblock Ordisco: Effective and efficient usage of incremental unlabeled data for semi-supervised continual learning.
\newblock In {\em CVPR}, pages 5383--5392, 2021.

\bibitem[\protect\citeauthoryear{Wu \bgroup \em et al.\egroup }{2024}]{wu2024unsupervised}
Di~Wu, Shicai Fan, Xue Zhou, Li~Yu, Yuzhong Deng, Jianxiao Zou, and Baihong Lin.
\newblock Unsupervised anomaly detection via masked diffusion posterior sampling.
\newblock In {\em IJCAI}, pages 2442--2450, 2024.

\bibitem[\protect\citeauthoryear{Yang \bgroup \em et al.\egroup }{2025}]{yang2025defect}
Shuai Yang, Zhifei Chen, Pengguang Chen, Xi~Fang, Yixun Liang, Shu Liu, and Yingcong Chen.
\newblock Defect spectrum: a granular look of large-scale defect datasets with rich semantics.
\newblock In {\em ECCV}, pages 187--203. Springer, 2025.

\bibitem[\protect\citeauthoryear{Yoon \bgroup \em et al.\egroup }{2017}]{Architecture-3}
Jaehong Yoon, Eunho Yang, Jeongtae Lee, and Sung~Ju Hwang.
\newblock Lifelong learning with dynamically expandable networks.
\newblock {\em arXiv preprint arXiv:1708.01547}, 2017.

\bibitem[\protect\citeauthoryear{You \bgroup \em et al.\egroup }{2022}]{Reconstruction_Uniad}
Zhiyuan You, Lei Cui, Yujun Shen, Kai Yang, Xin Lu, Yu~Zheng, and Xinyi Le.
\newblock A unified model for multi-class anomaly detection.
\newblock {\em NIPS}, 35:4571--4584, 2022.

\bibitem[\protect\citeauthoryear{Zavrtanik \bgroup \em et al.\egroup }{2021a}]{synthesizing_draem}
Vitjan Zavrtanik, Matej Kristan, and Danijel Sko{\v{c}}aj.
\newblock Draem-a discriminatively trained reconstruction embedding for surface anomaly detection.
\newblock In {\em ICCV}, pages 8330--8339, 2021.

\bibitem[\protect\citeauthoryear{Zavrtanik \bgroup \em et al.\egroup }{2021b}]{AP}
Vitjan Zavrtanik, Matej Kristan, and Danijel Sko{\v{c}}aj.
\newblock Draem-a discriminatively trained reconstruction embedding for surface anomaly detection.
\newblock In {\em ICCV}, pages 8330--8339, 2021.

\bibitem[\protect\citeauthoryear{Zhang \bgroup \em et al.\egroup }{2024a}]{Reconstruction_Invad}
Jiangning Zhang, Chengjie Wang, Xiangtai Li, Guanzhong Tian, Zhucun Xue, Yong Liu, Guansong Pang, and Dacheng Tao.
\newblock Learning feature inversion for multi-class anomaly detection under general-purpose coco-ad benchmark.
\newblock {\em arXiv preprint arXiv:2404.10760}, 2024.

\bibitem[\protect\citeauthoryear{Zhang \bgroup \em et al.\egroup }{2024b}]{Invad}
Jiangning Zhang, Chengjie Wang, Xiangtai Li, Guanzhong Tian, Zhucun Xue, Yong Liu, Guansong Pang, and Dacheng Tao.
\newblock Learning feature inversion for multi-class anomaly detection under general-purpose coco-ad benchmark.
\newblock {\em arXiv preprint arXiv:2404.10760}, 2024.

\bibitem[\protect\citeauthoryear{Zhou and Paffenroth}{2017}]{AUROC}
Chong Zhou and Randy~C Paffenroth.
\newblock Anomaly detection with robust deep autoencoders.
\newblock In {\em SIGKDD}, pages 665--674, 2017.

\bibitem[\protect\citeauthoryear{Zou \bgroup \em et al.\egroup }{2022}]{Visa}
Yang Zou, Jongheon Jeong, Latha Pemula, Dongqing Zhang, and Onkar Dabeer.
\newblock Spot-the-difference self-supervised pre-training for anomaly detection and segmentation.
\newblock In {\em ECCV}, pages 392--408. Springer, 2022.

\bibitem[\protect\citeauthoryear{Zuo \bgroup \em et al.\egroup }{2024}]{zuo2024clip}
Zuo Zuo, Yao Wu, Baoqiang Li, Jiahao Dong, You Zhou, Lei Zhou, Yanyun Qu, and Zongze Wu.
\newblock Clip-fsac: Boosting clip for few-shot anomaly classification with synthetic anomalies.
\newblock In {\em IJCAI}, pages 1834--1842, 2024.

\end{thebibliography}


\clearpage

\appendix
\section{Appendix}

\begin{figure}[h]
\centerline{\epsfig{figure=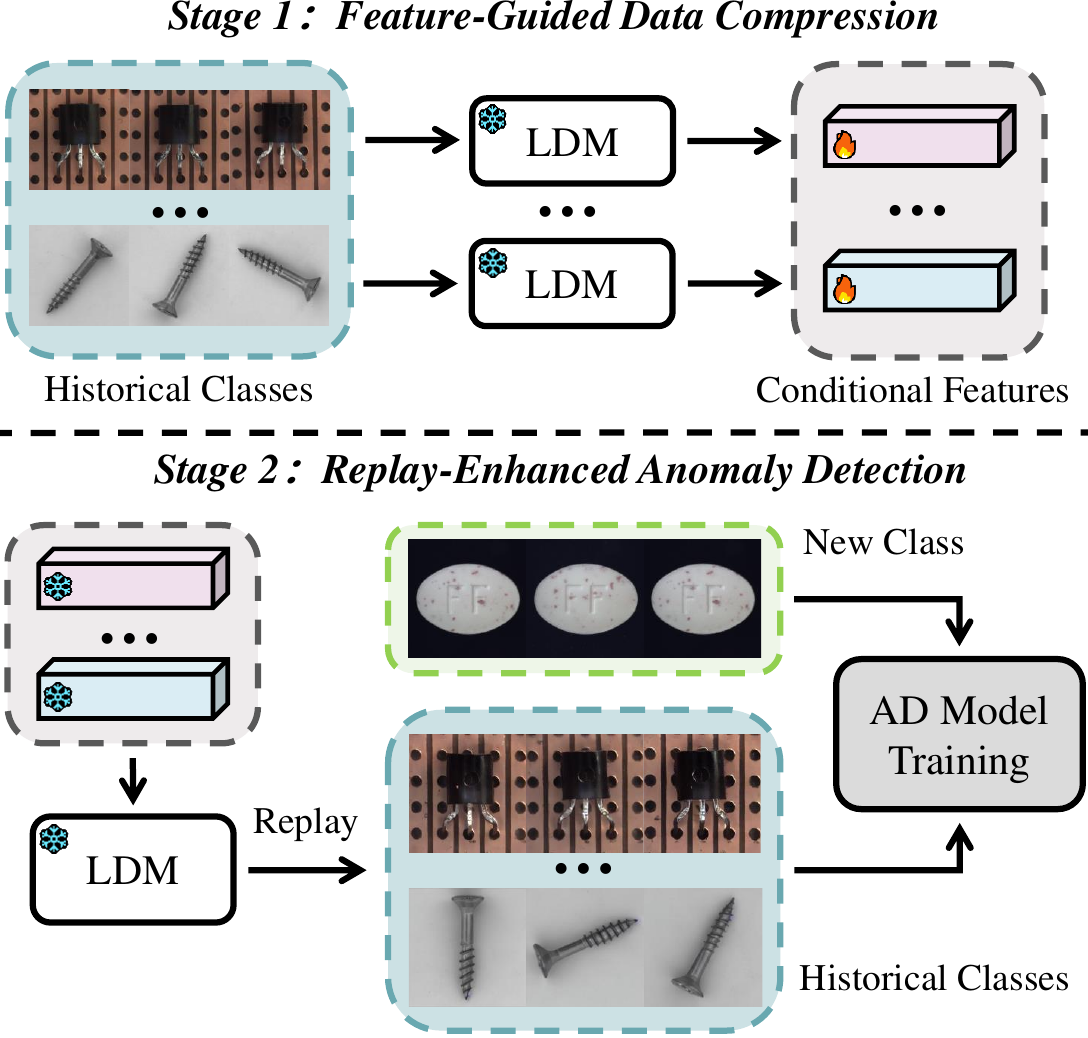,width=8cm}}
\caption{Overview of the ReplayCAD framework. In the first stage, the data of each historical class is compressed into conditional features and stored. In the second stage, the stored conditional features are used to replay the historical class data, which is then combined with new class data to train the AD model.}
\label{fig:12}
\end{figure}

This supplementary material consists of:

\begin{itemize}
  \item In Section \ref{sec:a1}, we have provided more ablation experiments.

  \item In Section \ref{sec:a2}, we provide image-level and pixel-level evaluation metrics for each class.

  \item In Section \ref{sec:a3}, we explain how to use SAM to obtain the mask for each sample.

  \item In Section \ref{sec:a4}, we present additional replayed data.

  \item In Figure \ref{fig:12}, we present the specific process of our approach.

  
\end{itemize}

\subsection{More ablation experiments}
\label{sec:a1}

\subsubsection{Qualitative analysis of semantic and spatial features.}
\begin{figure}[h]
\centerline{\epsfig{figure=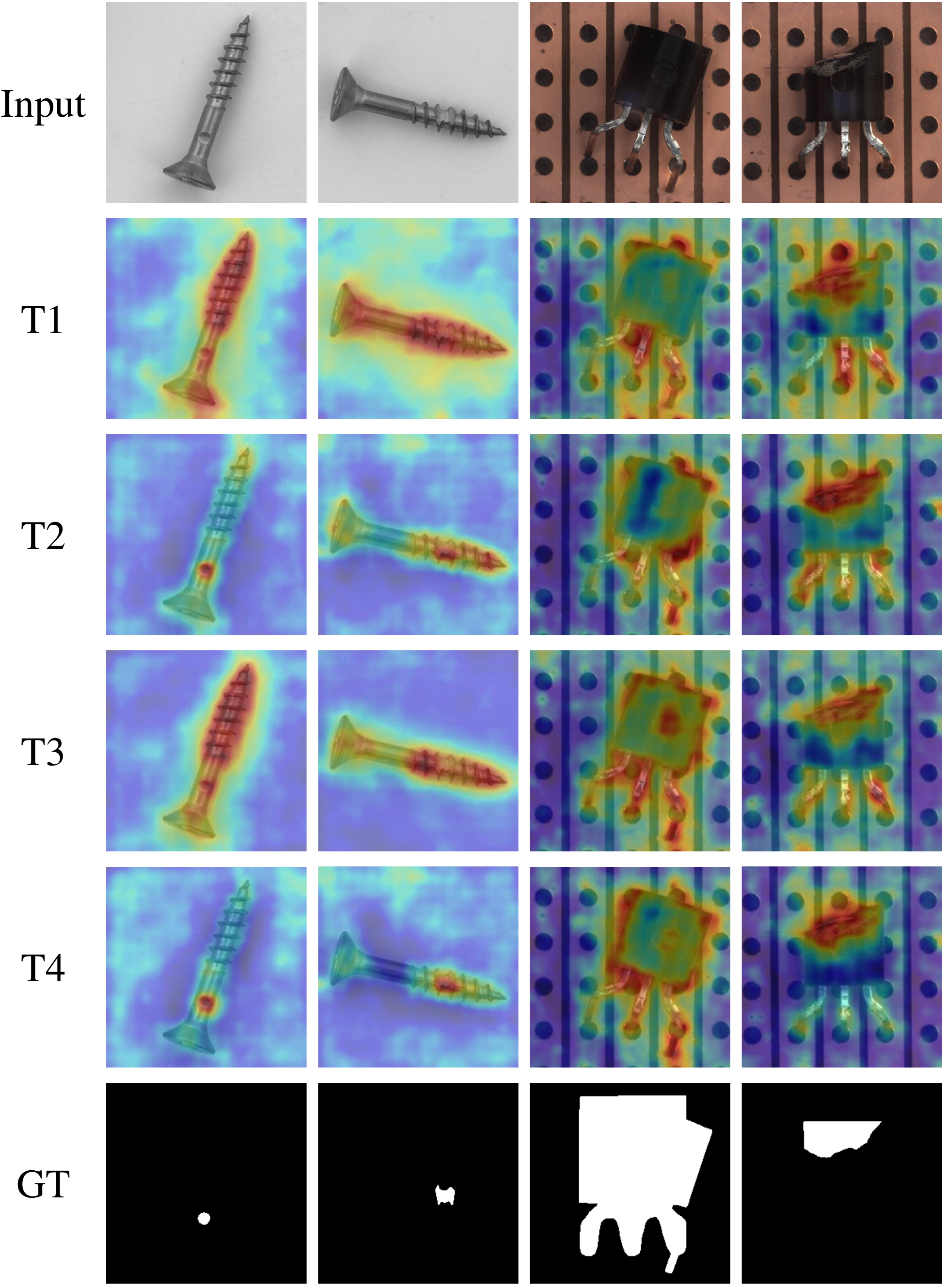,width=8cm}}
\caption{Qualitative analysis of semantic and spatial features in anomaly region segmentation. T1 means that neither feature is applicable, T2 and T3 mean that only semantic and spatial features are used respectively, and T4 means that both features are used simultaneously. Areas with high anomaly scores are shown in red, while areas with low anomaly scores are shown in blue.}
\label{fig:7}
\end{figure}

In Table \ref{tab:3}, we performed a quantitative analysis of the impact of semantic and spatial features. In this section, we further conducted a visual analysis to examine the impact of these two features on anomaly region segmentation. As shown in Figure \ref{fig:7}, when feature compression is not guided by semantic or spatial features, the model struggles to retain detailed information from past samples, resulting in poor segmentation performance. Incorporating either semantic or spatial features significantly improves segmentation, demonstrating their effectiveness. Ultimately, considering both semantic and spatial features leads to better segmentation results.

\begin{figure}[h]
\centerline{\epsfig{figure=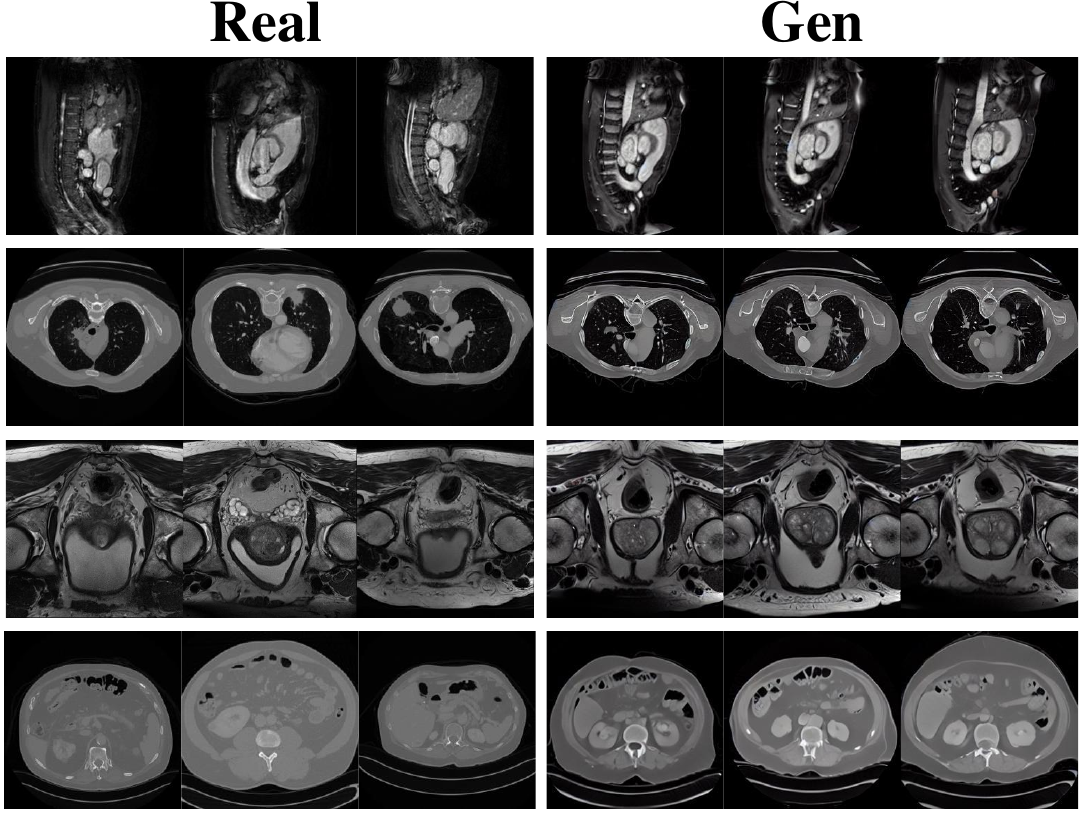,width=8cm}}
\caption{The generation results on MSD and kits23.}
\label{fig:r1}
\end{figure}

\subsubsection{Generalization.}

 We demonstrate the generalization capability of ReplayCAD in the following two points: 1)  We validated ReplayCAD on two widely used medical image datasets: MSD~\cite{MSD} and kits23~\cite{kits}. As shown in Figure \ref{fig:r1}, the results below show that ReplayCAD can effectively generate data even when there are significant differences from the pretraining domain. 2) The MVTec and VisA datasets we used were not used for pretraining the diffusion model.

 \begin{table}[h]
 \centering
  \renewcommand\arraystretch{1.2}
  \tabcolsep=0.22cm
 \caption{Effect of different spatial information on MVTec.}
\label{tab:r2}
\begin{tabular}{c|cccc}
\hline
 Method & Mask & Noisy mask& MBR&  \\ \hline
  Performance& \textbf{94.8/53.7} & 93.8/52.1 & 94.4/52.8
 &  \\\hline
\end{tabular}
\end{table}

 \begin{figure}[H]
\centerline{\epsfig{figure=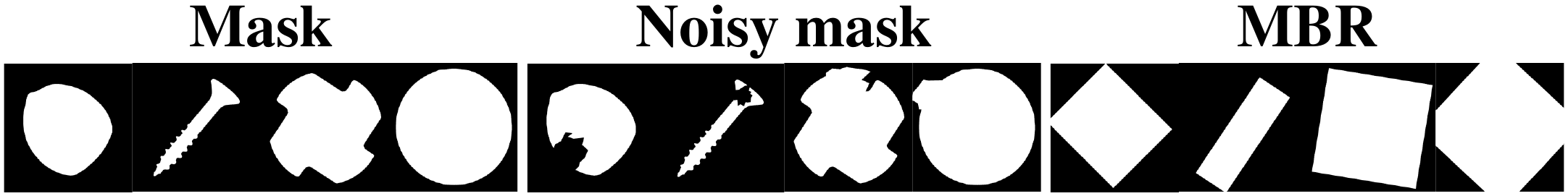,width=8.5cm}}
 \caption{Different spatial information.}
\label{fig:r2}
\end{figure}

\subsubsection{Ablation experiment of spatial information.}

As in Table~\ref{tab:r2} and Figure \ref{fig:r2}, We are robust to SAM's segmentation errors: First, we add random noise (Increase or decrease the segmentation area) to 30\% of training and inference masks. Second, we directly use the Minimum Bounding Rectangle (MBR) of masks. Experiments prove the model is robust to noisy spatial information.

\subsubsection{Ablation experiment of the number of semantic embedding tokens (K).}

We set K=20 in both MVTec and VisA as it represents the optimal balance point between performance and storage space. As shown in Table~\ref{tab:r3}, a small K (1 or 5) fails to sufficiently preserve sample features, leading to poorer performance, while a large K (40) significantly increases storage space while having minimal impact on performance.

\begin{table}[h]
 \centering
  \renewcommand\arraystretch{1.2}
 \tabcolsep=0.12cm
  \caption{Ablation experiment of K on MVTec.}
\label{tab:r3}
\begin{tabular}{c|ccccc}
\hline
 K & 1 & 5 & 20 & 40 \\ \hline
 Storage space & \textbf{1.5 MB} & 1.8 MB & 3.0 MB & 4.5 MB \\ 
Performance & 90.3/47.5 & 92.6/49.4& \textbf{94.8/53.7}& 94.6/53.5 \\ \hline
\end{tabular}
\end{table}

\subsubsection{Ablation experiment of class order.}

As shown in Table~\ref{tab:r1}, we train the model using four random class orders and the performances are close, demonstrating that class order does not affect model performance. 
\begin{table}[h]
 \centering
 \renewcommand\arraystretch{1.2}
  \tabcolsep=0.11cm
 \caption{Effect of class order on MVTec.}

\label{tab:r1}
\begin{tabular}{c|ccccc}
\hline
 Class order & 1 & 2 & 3
&4 &  \\ \hline
 Performance& 94.8/53.7 & 94.1/52.1
 &95.3/52.2 & 95.5/52.7&  \\ \hline
\end{tabular}
\end{table}

\subsection{Quantitative results on each class}
\label{sec:a2}
In Table \ref{tab:a1}, we list the image-level and pixel-level evaluation metrics for each class on the VisA dataset. We achieved the best or second-best performance across all classes, particularly excelling in segmentation for classes like \textit{Candle}, \textit{Macaroni1}, \textit{Macaroni2}, \textit{Fryum}, and \textit{Pcb4}. This is because these classes have rich detailed features, and our method is better at preserving these features, leading to superior segmentation results. As shown in Table \ref{tab:a2}, we present the results on the MVTec dataset, our method demonstrates significant advantages, further confirming its effectiveness.

\subsection{Mask Generation}
\label{sec:a3}

\begin{figure}[h]
\centerline{\epsfig{figure=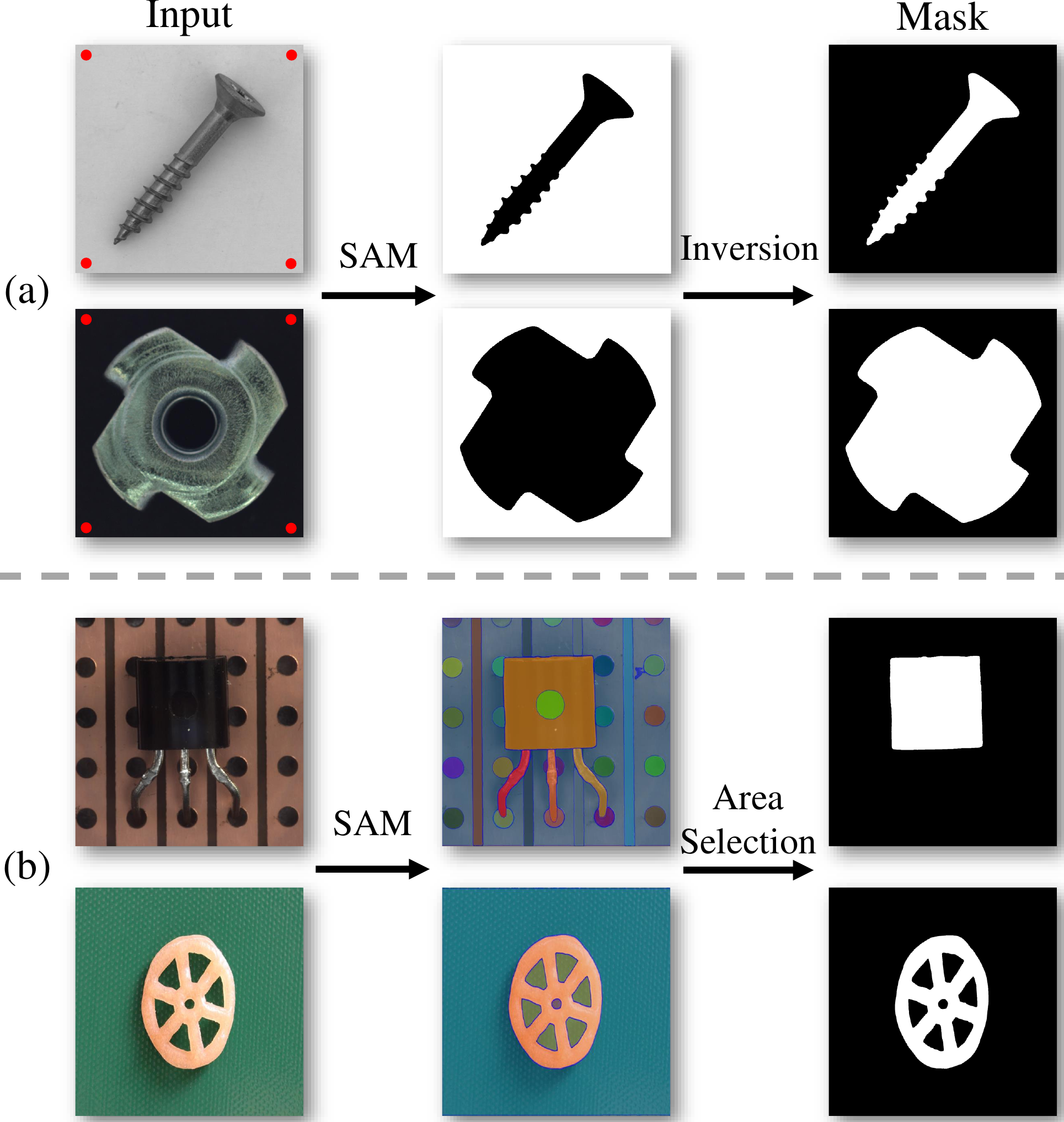,width=8.5cm}}
\caption{Two methods for generating masks. (a) Points prompt. We first use the four corners of the image as prompts to obtain the background mask, then apply an inversion to get the sample mask. (b) Area selection. We directly input the original image into SAM to obtain masks for all targets, then filter the target sample masks based on area.}
\label{fig:8}
\end{figure}

We employ two methods, points prompt and area selection, to obtain the sample mask. As shown in Figure \ref{fig:8} (a), for samples with simple and unobstructed backgrounds (such as screws and metal nuts), we use points prompt to acquire the masks. Specifically, we input the coordinates of the four corners of the image as prompts into SAM along with the image to obtain the background mask area, with the remaining part being the sample mask area. As illustrated in Figure \ref{fig:8} (b), for samples with complex backgrounds (such as transistors and fryums), we use area selection to obtain the masks. Specifically, we first input the image directly into SAM to acquire masks for all objects. Considering that the area of each type of sample is relatively fixed, we directly select the mask with the corresponding area size as the sample mask.

\subsection{Additional replayed data}
\label{sec:a4}

\begin{figure*}[]
\centerline{\epsfig{figure=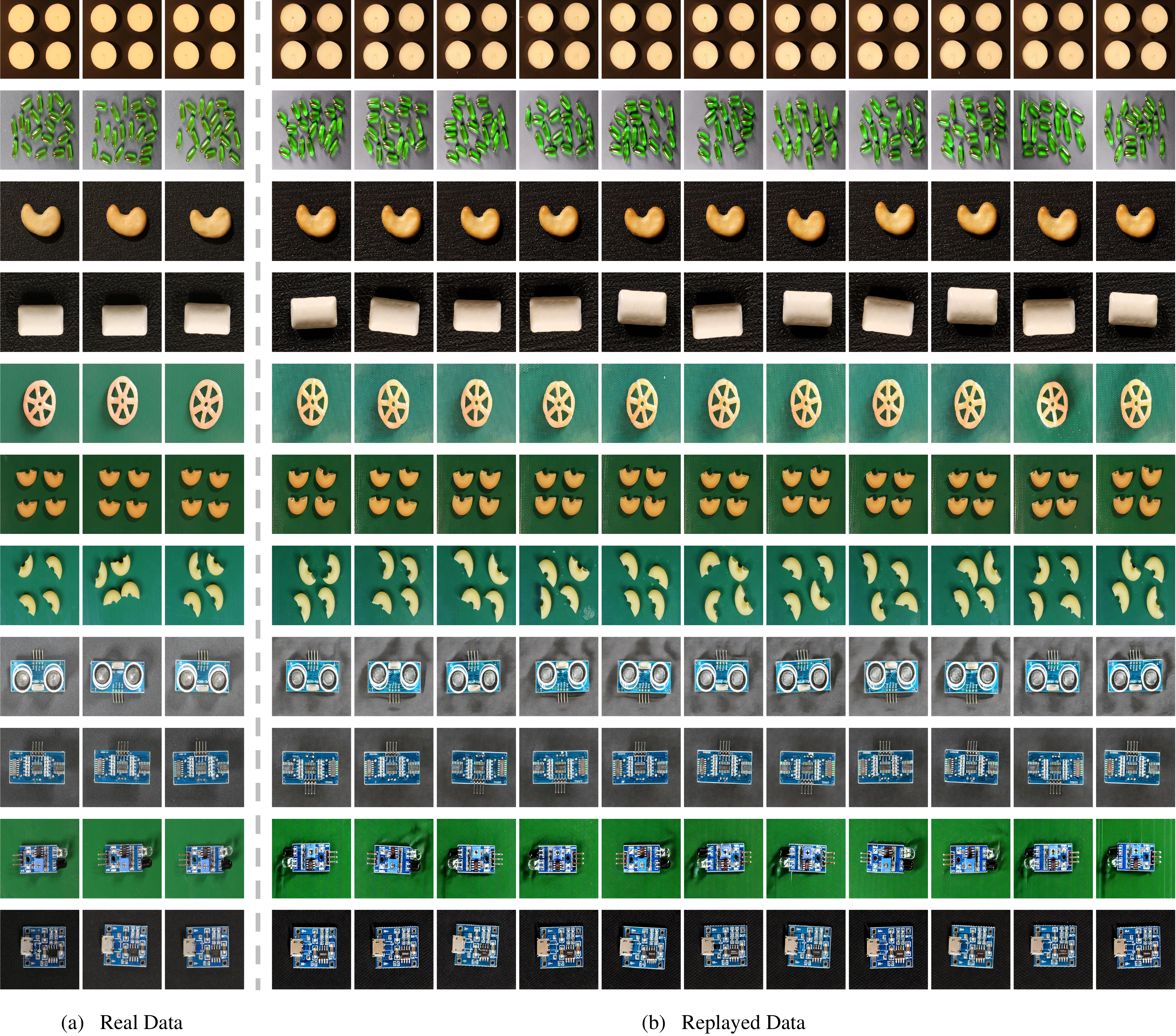,width=17.5cm}}
\caption{The visual comparison of replayed data and real data of each class on VisA. The classes from top to bottom are \textit{Candle}, \textit{Capsules}, \textit{Cashew}, \textit{Chewinggum}, \textit{Fryum}, \textit{Macaroni1}, \textit{Macaroni2}, \textit{Pcb1}, \textit{Pcb2}, \textit{Pcb3} and \textit{Pcb4}.}
\label{fig:10}
\end{figure*}

\begin{figure*}[]
\centerline{\epsfig{figure=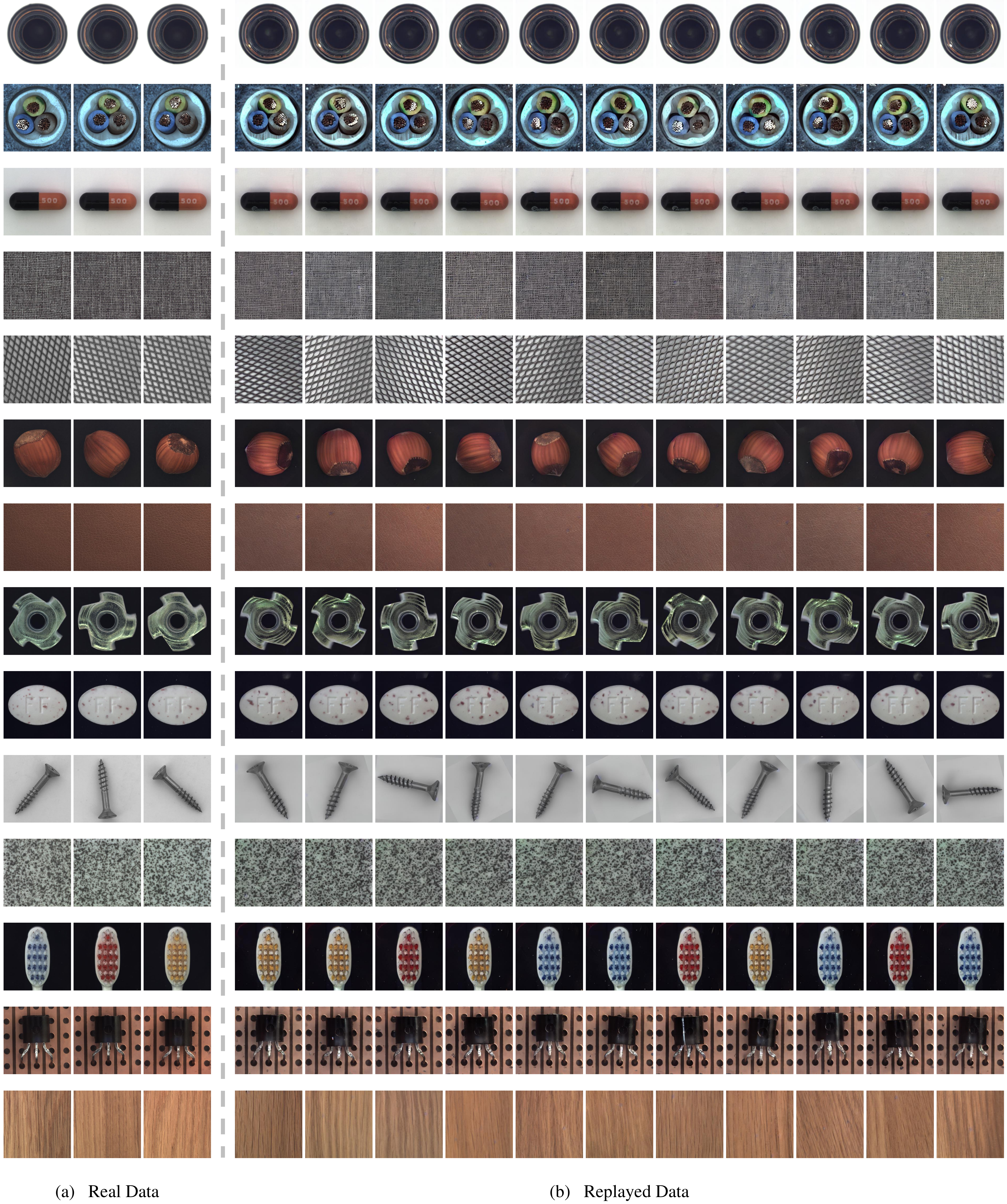,width=17.5cm}}
\caption{The visual comparison of replayed data and real data of each class on MVTec. The classes from top to bottom are \textit{Bottle}, \textit{Cable}, \textit{Capsule}, \textit{Carpet}, \textit{Grid}, \textit{Hazelnut}, \textit{Leather}, \textit{Metal Nut}, \textit{Pill}, \textit{Screw}, \textit{Tile}, \textit{Toothbrush}, \textit{Transistor} and \textit{Wood}.}
\label{fig:9}
\end{figure*}

As shown in Figure \ref{fig:10}, \ref{fig:9}, we provide additional replayed data on the VisA and MVTec datasets. We can see that the replayed data effectively restore the detailed semantic features of the real data, and they exhibit diverse spatial positions, demonstrating the effectiveness of our framework.

\begin{table*}[]
\begin{center}
\tabcolsep=0.2cm
\renewcommand\arraystretch{1.5}
\caption{Quantitative results (Image-AUROC / Pixel-AP) for each class on VisA. $* $ Indicates that MAS is adapted. \textbf{Bold} indicates the best performance, while \underline{underline} indicates the second-best performance.}
\label{tab:a1}
\begin{tabular}{c|>{\centering\arraybackslash}p{1.4cm}>{\centering\arraybackslash}p{1.4cm}>{\centering\arraybackslash}p{1.4cm}>{\centering\arraybackslash}p{1.4cm}>{\centering\arraybackslash}p{1.4cm}>{\centering\arraybackslash}p{1.4cm}>{\centering\arraybackslash}p{1.4cm}>{\centering\arraybackslash}p{1.4cm}>{\centering\arraybackslash}p{1.4cm}}
\specialrule{1.5pt}{0pt}{0pt}
\multirow{2}{*}{Class} & \multicolumn{9}{c}{Method} \\ \cline{2-10} 
 & PatchCore & SimpleNet & MambaAD   & {SimpleNet}$^{*}$& MambaAD$^{*}$ & DNE & IUF  & UCAD & ReplayCAD  \\ \hline
 
 \textit{Candle}& 39.7/1.1 & 25.3/0.3 & 59.5/0.6 & 42.0/0.9 &  62.4/0.6 & 21.9/-- & {\bf99.4}/1.2 & 77.8/\underline{6.7} & \underline{92.4}/{\bf24.1} \\ 
 
 \textit{Capsules}& 65.1/0.7 & 52.0/0.6 & 40.2/0.8 & 60.6/0.6 &  49.6/1.1  & 31.9/-- & 69.2/1.7 & {\bf87.7}/{\bf43.7} & \underline{84.3}/\underline{43.0} \\
 
 \textit{Cashew}& 72.1/6.2 & 85.3/6.1 & 81.2/14.9 &  73.3/12.1  & 75.4/5.0 & 81.5/-- & 75.8/4.3 & {\bf96.0}/{\bf58.0} & \underline{93.7}/\underline{55.5} \\
 
 \textit{Chewinggum}& 92.4/39.2 & 55.2/0.4 & 73.7/23.8 & 49.2/0.6 & 89.6/37.6 & 76.3/-- & 54.8/3.3 & \underline{95.8}/\underline{50.3} & {\bf96.1}/{\bf67.4} \\
 
 \textit{Fryum}& 42.3/10.0 & 35.0/1.7 & 48.4/10.4 & 53.7/5.8 &  56.4/10.5  & 83.2/-- & 67.7/10.7 & {\bf94.5}/\underline{33.4} & \underline{91.5}/{\bf46.2} \\
 
 \textit{Macaroni1}& 46.0/0.1 & 54.1/0.1 & 57.2/0.3 &  55.8/0.1  & 59.6/0.2 & 39.9/-- & 79.5/1.1 & \underline{82.3}/\underline{1.3} & {\bf88.9}/{\bf17.8} \\
 
 \textit{Macaroni2}& 33.8/0.1 & 59.0/0.1 & 42.8/0.1 &  27.3/0.1  & 51.7/0.1 & 29.9/-- & 60.6/\underline{0.4} & \underline{66.7}/0.3 & {\bf80.5}/{\bf9.9} \\
 
 \textit{Pcb1}& 61.2/1.1 & 74.9/7.0 & 57.3/1.1 &  72.2/13.8  & 58.8/1.6 & 76.9/-- & 56.3/1.9 & \underline{90.5}/\underline{70.2} & {\bf91.1}/{\bf79.3} \\
 
 \textit{Pcb2}& 59.5/0.8 & 64.0/0.2 & 55.6/0.7 & 56.8/0.8 & 57.6/1.2 &  54.5/--  & 76.6/0.9 & {\bf87.1}/\underline{13.6} & \underline{84.9}/{\bf19.9} \\
 
 \textit{Pcb3}& 42.1/0.9 & 60.6/0.2 & 49.2/1.0 & 65.9/0.2 & 46.4/1.3 &  58.6/--  & 65.1/1.8 & \underline{81.3}/\underline{26.6} & {\bf83.1}/{\bf42.2} \\
 
 \textit{Pcb4}& 37.0/1.3 & 55.7/1.7 & 50.7/1.8 & 79.0/9.6 & 77.4/5.6 &  75.0/--  & 51.2/2.1 & \underline{90.1}/\underline{10.6} & {\bf97.8}/{\bf30.3} \\
 
 \textit{Pipe Fryum}& {\bf 99.8}/62.8 & 86.4/{\bf70.0} & 98.4/50.8 & 91.9/\underline{67.4} & 98.2/49.2 & 92.5/-- & 61.4/11.7 & 98.8/45.7 & \underline{99.1}/62.5 \\ \hline
 
 Avg& 57.6/10.3 &58.9/7.3 & 59.5/8.8 & 60.6/9.3 & 65.2/9.5 & 60.1/-- & 68.1/3.4 & \underline{87.4}/\underline{30.0} & {\bf90.3}/{\bf41.5} \\\specialrule{1.5pt}{0pt}{0pt}
\end{tabular}
\end{center}
\end{table*}

\begin{table*}[]
\begin{center}
\tabcolsep=0.2cm
\renewcommand\arraystretch{1.5}
\caption{Quantitative results (Image-AUROC / Pixel-AP) for each class on MVTec. $* $ Indicates that MAS is adapted. \textbf{Bold} indicates the best performance, while \underline{underline} indicates the second-best performance.}
\label{tab:a2}
\begin{tabular}{c|>{\centering\arraybackslash}p{1.4cm}>{\centering\arraybackslash}p{1.4cm}>{\centering\arraybackslash}p{1.4cm}>{\centering\arraybackslash}p{1.4cm}>{\centering\arraybackslash}p{1.4cm}>{\centering\arraybackslash}p{1.4cm}>{\centering\arraybackslash}p{1.4cm}>{\centering\arraybackslash}p{1.4cm}>{\centering\arraybackslash}p{1.4cm}}
\specialrule{1.5pt}{0pt}{0pt}
\multirow{2}{*}{Class} & \multicolumn{9}{c}{Method} \\ \cline{2-10} 
 & PatchCore & SimpleNet & MambaAD   & {SimpleNet}$^{*}$& MambaAD$^{*}$ & DNE & IUF  & UCAD & ReplayCAD  \\ \hline
 
 \textit{Bottle}& 21.3/5.2 & 82.3/36.9 & 79.2/18.7 & 65.9/28.4 &  85.3/15.0 & 97.9/-- & 90.9/28.9 & {\bf 100.0}/{\bf75.2} & \underline{99.0}/\underline{71.0}\\ 
 
 \textit{Cable}& 51.0/3.4 & 46.1/4.5 & 45.1/4.5 & 59.5/18.7 &  43.8/4.2  & 74.4/-- & 54.1/5.4 & \underline{75.1}/\underline{29.0} & {\bf 95.7}/{\bf36.9} \\
 
 \textit{Capsule}& 34.2/4.3 & 55.3/3.3 & 41.4/5.6 &  38.0/3.1 & 52.3/9.4 & 57.0/-- & 52.0/4.0 & {\bf 86.8}/{\bf34.9} & \underline{74.7}/\underline{33.7} \\
 
 \textit{Carpet}& 97.0/60.0 & 75.8/10.7 & 67.6/11.7 &  87.4/40.7 & 61.3/9.7 & {\bf 99.6}/-- & {\bf 99.6}/44.0 & 96.5/\underline{62.2} & \underline{98.0}/{\bf65.2} \\
 
 \textit{Grid}& 70.6/0.5 & 49.7/0.8 & 55.8/0.6 & 65.6/0.7 &  54.5/0.6  & {\bf 98.3}/-- & 69.5/8.4 & \underline{94.4}/\underline{18.7} & 92.7/{\bf33.8} \\
 
 \textit{Hazelnut}& 84.8/38.4 & 37.4/5.6 & 83.4/30.0 &  72.6/21.5  & 85.5/30.9 & 94.8/-- & 87.5/30.1 & {\bf 99.4}/\underline{50.6} & \underline{98.5}/{\bf63.5} \\
 
 \textit{Leather}& 61.8/29.7 & 43.7/0.5 & 53.2/8.8 &  75.3/15.7  & 53.0/5.2 & 88.6/-- & \underline{99.7}/33.0 & {\bf 100.0}/\underline{33.3} & 97.4/{\bf 58.7} \\
 
 \textit{Metal Nut}& 26.7/25.7 & 52.2/42.6 & 40.7/19.4 & 50.7/49.2 & 60.5/17.5 & 98.1/-- & 64.3/14.2 & \underline{98.8}/{\bf 77.5} & {\bf99.5}/\underline{65.6} \\
 
 \textit{Pill}& 45.4/5.8 & 62.1/19.6 & 46.8/5.5 & 54.7/22.1 & 54.2/4.8 &  80.6/--  & 54.7/4.8 & \underline{89.4}/\underline{63.4} & {\bf 94.4}/{\bf69.8} \\
 
 \textit{Screw}& 49.1/1.1 & 57.7/1.3 & 52.0/1.2 & 60.8/1.5 & 53.9/1.0 &  61.8/--  & 64.6/1.2 & \underline{73.9}/\underline{21.4} & {\bf 79.5}/{\bf32.9} \\
 
 \textit{Tile}& 77.4/28.1 & 94.7/\underline{59.1} & 75.9/9.3 & 94.4/{\bf66.7} & 75.6/9.6 &  96.8/--  & 94.0/31.0 & \underline{99.8}/54.9 & {\bf 99.9}/53.1 \\
 
 \textit{Toothbrush}& 60.3/4.7 & 68.9/29.2 & 49.2/3.6 & 80.8/\underline{ 31.6 } & 41.7/4.7 & 92.2/-- & 71.1/4.9 & {\bf 100.0}/29.8 & \underline{98.1}/{\bf57.6} \\
 
 \textit{Transistor}& 34.6/8.3 & 53.1/4.6 & 47.2/5.6 & 63.6/13.4 & 66.3/7.6 & 84.4/-- & 66.0/6.5 & \underline{87.4}/\underline{39.8} & {\bf 95.7}/{\bf60.5} \\
 
 \textit{Wood}& 97.0/34.1 & 58.8/3.8 & 83.0/15.6 & 86.5/20.4 & 90.1/15.0 &  94.4/--  & 95.3/32.6 & {\bf 99.5}/{\bf53.5} & \underline{98.4}/\underline{50.0} \\
 
 \textit{Zipper}& \underline{99.4}/{\bf63.9} & 98.0/43.9 & 89.3/50.4 & 99.3/46.9 &  89.3/48.0  & 95.5/-- & 79.5/8.0 & 93.8/39.8 & {\bf 99.7}/\underline{53.9} \\ \hline
 
 {\bf Avg}& 60.7/20.9 & 62.4/17.7 & 60.6/12.7 & 70.3/25.4 & 64.5/12.2 & 87.6/-- & 76.2/17.1 & \underline{93.0}/\underline{45.6} & {\bf94.8}/{\bf53.7} \\\specialrule{1.5pt}{0pt}{0pt}
\end{tabular}
\end{center}
\end{table*}

\end{document}